\documentclass[10pt]{article} % For LaTeX2e
\usepackage[preprint]{tmlr}
\usepackage{amsmath,amsfonts,bm}

\def\eqref#1{equation~\ref{#1}}
\def\1{\bm{1}}

\DeclareMathAlphabet{\mathsfit}{\encodingdefault}{\sfdefault}{m}{sl}
\SetMathAlphabet{\mathsfit}{bold}{\encodingdefault}{\sfdefault}{bx}{n}

\usepackage{hyperref}
\usepackage{url}
\usepackage{pgffor}
\usepackage{caption}
\usepackage{graphicx}
\usepackage{wrapfig}
\usepackage{enumitem}
\usepackage[table,xcdraw]{xcolor}
\usepackage{booktabs}
\usepackage{nicematrix}
\usepackage{subcaption}
\usepackage[ruled,vlined]{algorithm2e}

\newcommand{\method}{Geospatial Diffusion-based Evolution Synthesis }
\newcommand{\arc}{GeoDES}

\title{Geospatial Diffusion-based Evolution Synthesis (GeoDES) for Storm-Centered Weather Augmentation}

\author{\name Sonia Cromp \email sonic@cs.wisc.edu \\
      \addr University of Wisconsin-Madison \vspace{-0.3cm}
      \AND
      \name Satya Sai Srinath Namburi GNVV \email sgnamburi@wisc.edu \\
      \addr GE HealthCare \vspace{-0.3cm}
      \AND
      \name Youran Wang \email wang3488@wisc.edu\\
      \addr University of Wisconsin-Madison \vspace{-0.3cm}
      \AND
      \name Grace Kisslinger \email grace.kisslinger@uga.edu \\
      \addr University of Georgia \vspace{-0.3cm}
      \AND
      \name Frederic Sala \email fredsala@cs.wisc.edu \\
      \addr University of Wisconsin-Madison \vspace{-0.3cm}
      \AND
      \name James Booth \email jbooth@ccny.cuny.edu \\
      \addr City College of New York \vspace{-0.3cm}
      \AND
      \name Allegra LeGrande \email al2035@columbia.edu \\
      \addr NASA Goddard Institute for Space Studies \vspace{-0.3cm}
      }

\def\month{MM}  % Insert correct month for camera-ready version
\def\year{YYYY} % Insert correct year for camera-ready version
\def\openreview{\url{https://openreview.net/forum?id=XXXX}} % Insert correct link to OpenReview for camera-ready version

\begin{document}

\maketitle

\begin{abstract} 
\vspace{-0.4cm}
While machine learning-based weather models hold significant promise, they struggle to predict the detailed structure of large-scale weather systems such as cyclonic storms. 
Regional models are constrained by limited historical records within fixed geographic boundaries, while global models are computationally expensive and often operate at resolutions too coarse to capture fine-grained storm dynamics. 
To bridge this gap, we introduce the Geospatial Diffusion-based Evolution Synthesis (GeoDES) model, 
a custom image-to-video diffusion model.
By focusing generation strictly on the evolving storm structure, GeoDES synthesizes physically consistent, high-fidelity weather events suitable for stress-testing forecast models and expanding meteorological datasets. 
Evaluations demonstrate that GeoDES outperforms prior methods on key metrics, achieving $52\%$ lower Peak Vorticity Error and $8\%$ higher Anomaly Correlation Coefficient than the next strongest methods on the North Atlantic test set.
\end{abstract}

\section{Introduction}
\vspace{-0.35cm}

The majority of weather hazards that impact Europe and North America, from flooding to straight-line winds, are associated with \textit{extratropical cyclone} storms. While most of these storms are benign, the rare extreme instances are notoriously difficult to predict. Because they lack the simplifying properties of other storms (e.g., hurricanes' axisymmetry \citep{willoughby2006parametric}), reliable probabilistic models for these dangerous storms remain underdeveloped. Furthermore, extreme storms are inherently difficult to study due to limited observational sample sizes, which cause machine learning (ML) methods to underperform traditional physics-based forecasting on extreme events \citep{zhang2026physics}. As climate change shifts baseline weather conditions \citep{gertler2019changing}, the need for improved models of extratropical cyclones has become increasingly urgent for enabling the synthesis of realistic storm observations beyond past historical constraints. 

Preexisting ML weather models (Table~\ref{tab:baselines-summary}) encounter structural challenges when applied to extratropical cyclones. Current architectures remain largely constrained to either global or regional prediction \citep{shi2025deeplearningfoundationmodels}, both of which are poorly suited for tracking detailed, far-moving storms. Global models require immense compute---yielding heavily smoothed predictions at tractable resolutions or spurious dynamics via masking strategies---while regional models are restricted to static areas. Furthermore, both approaches waste significant training capacity on calm, non-stormy conditions.

We address these challenges by introducing \textbf{Geospatial Diffusion-based Evolution Synthesis (GeoDES)} for modeling extratropical cyclone structures. We follow a \textit{storm-centered} approach \citep[e.g.,][]{booth_extratropical_2018}, treating each cyclone as a ``video'' anchored to its moving low-pressure center (Figure \ref{fig:compare}). To support this, GeoDES incorporates architectural advancements tailored for atmospheric modeling. We treat time as an independent dimension to enable\textit{ non-autoregressive synthesis}, first training a 2D-image spatial model before using temporal inflation \citep{singer2022make} to create a 3D-video spatiotemporal model.

The storm-centered framing and temporal inflation strategy directly resolve structural limitations of existing models. By moving with the storm, GeoDES tracks detailed, far-moving systems without the constraints of static regional boundaries. Isolating the domain to the storm itself reduces wasted capacity spent modeling calm weather and sidesteps the immense compute and heavy smoothing associated with global grids. Further, by synthesizing the entire spatiotemporal storm lifecycle simultaneously, our non-autoregressive design avoids the compounding temporal errors of traditional, autoregressive methods. Additional benefits include:

\vspace{-0.2cm}
% \newpage
\begin{itemize}[topsep=0pt,itemsep=-0.08cm,leftmargin=*,partopsep=0pt, parsep=3pt]
    \item \textit{Downstream compatibility:} GeoDES' outputs are directly useful for meteorological analysis, where storm-centered data is a common format (e.g., \citealp{booth_extratropical_2018, cc1,cc4}).
\item \textit{High resolution:} By dedicating model capacity entirely to the storm area, GeoDES captures fine-scale dynamics without the smoothing typical of global models. Representing storms over a ($1600$km)$^2$ area in a $32\times32$ matrix corresponds to an effective resolution of up to $0.45^\circ\times0.45^\circ$, enabling detailed representation.
    \item \textit{Computational accessibility:} Because this localized strategy reduces the overall spatial domain, the memory footprint remains small, allowing for training from scratch on standard consumer-grade GPUs.
    % \item \textit{Stable, high-performance synthesis:} Variable-specific normalization explicitly stabilizes the generation of complex meteorological variables (Section \ref{sec:exp4}), while temporal inflation yields significant performance and efficiency gains (Section \ref{sec:exp3}).
\end{itemize}

\begin{figure}
    \centering
    \includegraphics[trim=103mm 20mm 95mm 20mm, clip, width=0.9\textwidth]{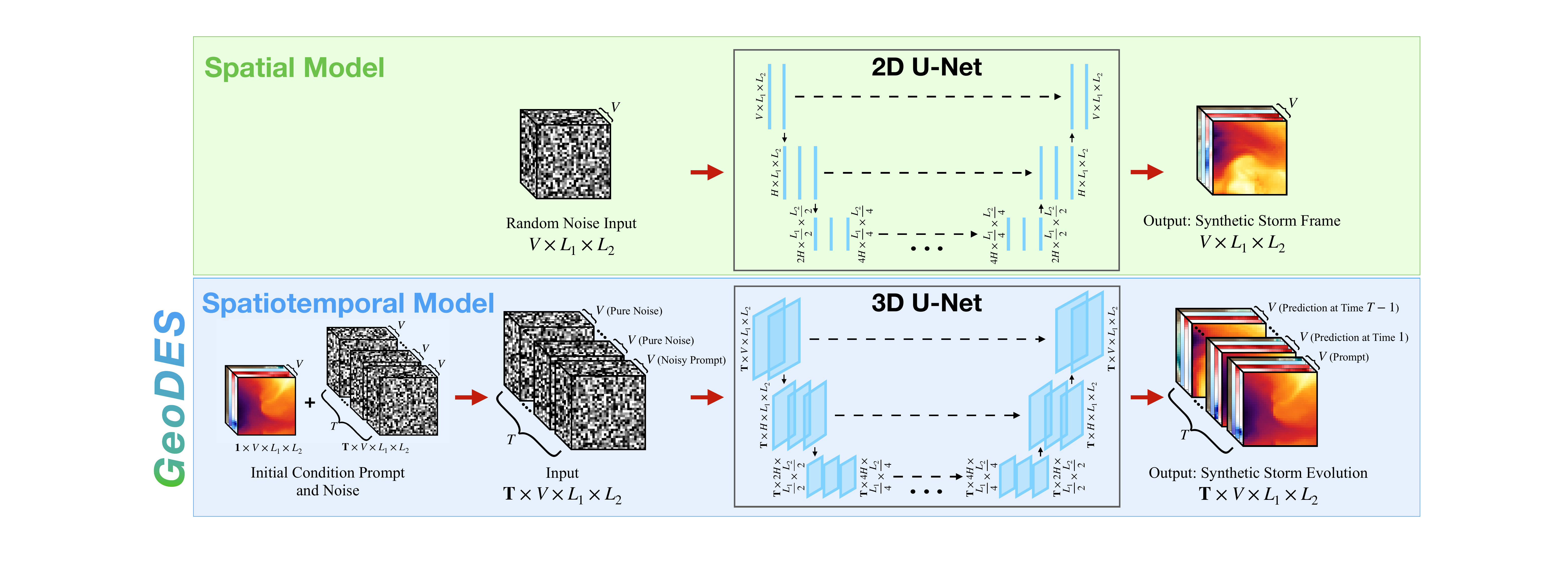}
    % \vspace{-0.5cm}
    \caption{Overview of the GeoDES model with hidden dimension $H$ (at bottom, with blue background), as well as the intermediate pretraining-phase model (at top, in green background), for a dataset of storm-centered bounding boxes with dimensions $L_1 \times L_2$, over $T$ timesteps and $V$ variables. GeoDES uses a prompt of initial storm conditions to predict the following $T-1$ timesteps of storm evolution.}
    \vspace{-0.45cm}
    \label{fig:flowchart}
\end{figure}

\begin{figure}
    % \vspace{-0.45cm}
    \centering
    \includegraphics[width=0.98\textwidth]{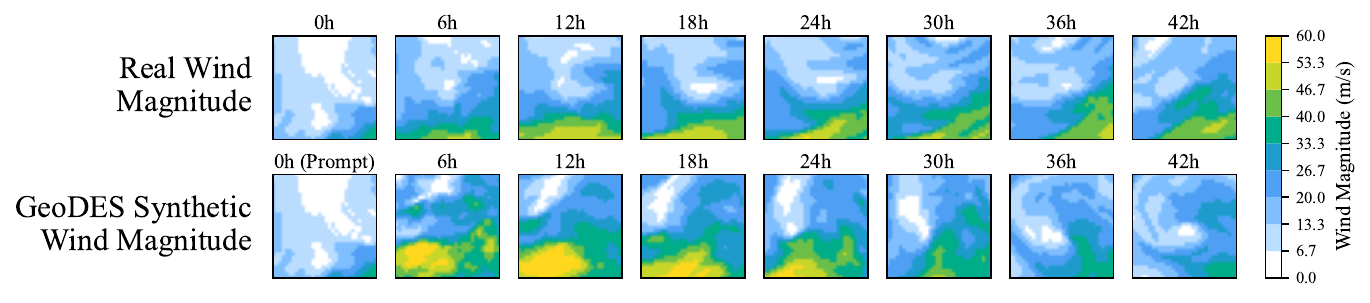}
    \vspace{-0.5cm}
    \caption{Prompted with real 0h wind magnitude data (500hPa), GeoDES synthesizes an alternative cyclone lifecycle (bottom) that maintains the complex, high-frequency patterns of the real storm (top) over 42 hours.}
    \vspace{-0.5cm}
    \label{fig:example-real-synth}
\end{figure}

Evaluated on North Atlantic extratropical cyclones from 2016-2024, \textbf{GeoDES outperforms existing methods across a variety of storm quality metrics}. As detailed in Section~\ref{sec:evals}, it achieves a 52\% lower Peak Vorticity Error and an 8\% higher Anomaly Correlation Coefficient than the next strongest baseline. Furthermore, GeoDES is uniquely capable of accurately recovering the kinetic energy distribution of ground-truth storms, achieving a near-perfect Frequency Bias Index of 1.02 and a High-Frequency Spectral Ratio of 0.95. This generative fidelity holds even for the top 10\% most extreme storms, allowing GeoDES to accurately synthesize severe weather phenomena without succumbing to the spatial smoothing or unphysical hallucinations common in global models. In summary, our core contributions are: 
\vspace{-0.1cm}
\begin{itemize}[topsep=0pt,itemsep=-0.08cm,leftmargin=*,partopsep=0pt, parsep=3pt]
    \item \textit{Task Formalization:} We formalize the data augmentation task of \textit{conditional storm synthesis}.
    \item \textit{Novel Architecture:} We present GeoDES, a spatiotemporal diffusion model that uses storm-centered framing and temporal inflation to bypass the computational limits and spatial smoothing of global models.
    \item \textit{State-of-the-Art Generative Realism:} We demonstrate that GeoDES significantly outperforms existing baselines (Table~\ref{tab:baselines-summary}) in preserving rotational structure and extreme wind energy distributions.
    \item \textit{Extreme Event Augmentation:} We show that GeoDES successfully generates physically consistent severe storms to enrich the training distribution of rare, high-intensity events for downstream climate study.
\end{itemize}

% Next, in Section~\ref{sec:prelim}, we provide background knowledge on diffusion models and extratropical cyclones. 
We formally define GeoDES in Section~\ref{sec:method}. Evaluations in Section~\ref{sec:evals} demonstrate GeoDES' generative fidelity compared to prior approaches. We then discuss prior work, future work and conclusions in Sections~\ref{sec:relatedwork}-\ref{sec:conclusion}.

% \newpage

% \vspace{-.4cm}
\section{The \method (\arc) Method} \label{sec:method}

We design GeoDES, a custom image-to-video diffusion model, to resolve the issues of prior ML models for extratropical cyclone synthesis. 
Rather than attempting to learn a complex spatiotemporal manifold from scratch---which leads to physical hallucinations or dynamic range collapse as demonstrated by our Section~\ref{sec:exp4} ablations---we decouple spatial structure from temporal evolution. GeoDES is constructed via a three-phase pipeline: (1) two-dimensional spatial pretraining, (2) temporal inflation and (3) three-dimensional spatiotemporal fine-tuning. This end-to-end procedure, formalized in Algorithm \ref{alg:geodes_training} and visualized in Figure~\ref{fig:compare}, serves as the foundation for the architectural details described below.

GeoDES is based on the Denoising Diffusion Probabilistic Model (DDPM) \citep{ho2020denoising}, which learns to generate synthetic data by reversing a Markovian process that progressively adds Gaussian noise to a true data sample $x_0$ over $N$ steps. Further background on diffusion models is provided in Appendix~\ref{app:dm-background} and a glossary of terms is available in Appendix~\ref{app:glossary}.

\RestyleAlgo{ruled}
\begin{algorithm}[]
\caption{GeoDES Two-Stage Training Procedure}
\label{alg:geodes_training} \small
\SetAlgoLined
\SetAlgoNoEnd
\DontPrintSemicolon

\SetKwInOut{Input}{Input}
\SetKwInOut{Output}{Output}
\SetKwComment{Comment}{$\triangleright$\ }{}
\SetKwProg{Fn}{Procedure}{:}{}

\Input{Spatiotemporal dataset $D_{\texttt{3D}}$, Uninitialized U-Nets $\Theta_2, \Theta_3$, Frame count $T$}
\Output{Trained 3D spatiotemporal model $\Theta_3^*$}
\BlankLine

\Fn{\textsf{Train}($\Theta$, $\mathcal{D}$, \textsf{NoiseStrategy})}{
    \While{not converged}{
        Sample batch from $\mathcal{D}$ and apply hybrid normalization \;
        Apply \textsf{NoiseStrategy} to form noisy batch \;
        Compute SNR-weighted denoising loss $\mathcal{L}$ \;
        Update weights: $\Theta \leftarrow \Theta - \eta \nabla_{\Theta} \mathcal{L}$ \;
    }
    \Return $\Theta$ \;
}
\BlankLine

\Comment{Phase 1: Two-Dimensional Spatial Pretraining}
\SetKw{Let}{let}
\Let $D_{\texttt{2D}}$ be $D_{\texttt{3D}}$ with dimension $T$ flattened into the sample dimension \;
$\Theta_2' \leftarrow \textsf{Train}(\Theta_2, D_{\texttt{2D}}, \text{Variance-preserving schedule})$ \;
\BlankLine

\Comment{Phase 2: Temporal Inflation (2D-to-3D Weight Conversion)}
\ForEach{parameter $w_2 \in \Theta_2'$ matching $w_3 \in \Theta_3$}{
    \lIf{$\mathrm{dim}(w_2) = \mathrm{dim}(w_3)$}{
        $w_3 \leftarrow w_2$
    }
    \lElse{
        $w_3 \leftarrow \frac{1}{T} \textsf{Repeat}(w_2, T)$ \Comment*[r]{Inflate: $(*) \to (T, *)$}
    }
}
\BlankLine

\Comment{Phase 3: Three-Dimensional Spatiotemporal Fine-Tuning}
$\Theta_3' \leftarrow \textsf{Train}(\Theta_3, D_{\texttt{3D}}, \text{Correlated noise } \boldsymbol{\epsilon}(t) \text{ conditioned on } \mathbf{x}_0(0))$ \;

\Return $\Theta_3^*$ \; 
\end{algorithm} 

% \vspace{-0.5cm}

\subsection{Phase 1: The Spatial Prior Network ($f_{\Theta_2}$)} \label{sec:method-train}

The goal of the first phase is to learn a Spatial Prior Network, denoted as $f_{\Theta_2}$. Weather systems are governed by strict spatial dependencies (e.g., the physical relationship between pressure gradients and surrounding wind fields). By training a two-dimensional image model first, $f_{\Theta_2}$ learns to accurately represent these multivariate atmospheric relationships without the compounding complexity of temporal dynamics.

Suppose each 2D training datapoint $\mathbf{x}_0 \in D_{\texttt{2D}}$ is a tensor $\mathbf{x}_0 \in \mathbb{R}^{V \times L_1 \times L_2}$, where $V$ denotes the number of climate variables (e.g., temperature, sea-level pressure) and $L_1$ and $L_2$ represent the spatial grid dimensions. To ensure $f_{\Theta_2}$ captures the full variability of the weather data without suffering dynamic range collapse, we first apply a variable-specific hybrid normalizer. We use log-scaling for long-tailed distributions (e.g., wind magnitude) and robust linear scaling for centered distributions (e.g., temperature). In both cases, each variable is scaled according to the $1$st and $99$th percentiles present in the training distribution.

The network $f_{\Theta_2}$ is parameterized by an unconditional pixel-space U-Net \citep{ronneberger2015u}. Beginning with randomly-initialized weights, we optimize the model using a variance-preserving schedule \citep{song2020score} and a Signal-to-Noise Ratio (SNR) weighted objective \citep{hang2023efficient},
$$ \mathcal{L}_{\text{2D}} = \mathbb{E}_{\mathbf{x}, \boldsymbol{\epsilon}, t} \left[ \lambda(t) \| \boldsymbol{\epsilon} - f_{\Theta_2}(\mathbf{x}_t, t) \|_2^2 \right], $$
where $\lambda(t)$ is the SNR weight. This weighting explicitly forces the model to dedicate representational capacity to intermediate noise scales---which correspond to meaningful, large-scale storm structures---rather than wasting capacity on imperceptible, high-frequency noise.

\vspace{-0.15cm}
\subsection{Phase 2: Temporal Inflation} \label{sec:method-inflate}
\vspace{-0.2cm}

Once the spatial priors are established, we transition to the Spatiotemporal Evolution Network, denoted as $f_{\Theta_3}$. To introduce time without destroying the atmospheric representations learned by $f_{\Theta_2}$, we perform temporal inflation (Phase 2 of Algorithm~\ref{alg:geodes_training}), adapting the strategy of \cite{singer2022make}.

Let $T$ denote the number of evolution timesteps. We initialize $f_{\Theta_3}$ by copying all spatially-equivalent blocks from $\Theta_2$ to $\Theta_3$. For any spatial block $w_2$ that requires an expanded temporal dimension in the 3D model, we initialize the corresponding weights $w_3$ at each time $t$ as $w_3(t)=w_2/T$.

\vspace{-0.15cm}
\subsection{Phase 3: The Spatiotemporal Evolution Network ($f_{\Theta_3}$)} \label{sec:method-model}
\vspace{-0.2cm}

With the architecture inflated, Phase 3 fine-tunes $f_{\Theta_3}$ on temporal dynamics. Suppose a 3D datapoint is a tensor $\mathbf{x}_0 \in \mathbb{R}^{T \times V \times L_1 \times L_2}$.

\textbf{Training Phase:} We use a standard forward diffusion process for training. 
Because atmospheric state is highly autocorrelated across time, we find using standard independent Gaussian noise for each frame disrupts this inherent consistency, allowing for physically impossible state jumps. While diffusion methods usually rely solely on network convolutions learning temporal smoothness, we explicitly inject this prior into the forward process by correlating the noise at frame $t$ with the preceding frame $t-1$ via a tunable parameter $\rho$.
Given uncorrelated noise $\boldsymbol{\epsilon}_u \sim \mathcal{N}(\mathbf{0}, \mathbf{I})$, we set $\boldsymbol{\epsilon}(0)=\boldsymbol{\epsilon}_u(0)$ and subsequent frames $t \in [1,T)$ as
$$ \boldsymbol{\epsilon}(t) = \rho\boldsymbol{\epsilon}(t-1) + \sqrt{1-\rho^2}\boldsymbol{\epsilon}_u(t). $$
In our experiments, we set $\rho = 0.95$. Real atmospheric variables exhibit complex, variable-specific covariance structures; thus, developing dynamic, physics-informed noise schedules remains a valuable direction for future work. The reverse process then trains $f_{\Theta_3}$ to denoise $\boldsymbol{\epsilon}$ using the same SNR-weighted objective as Phase 1.

\textbf{Sampling Phase:} At sample time, the goal is to predict future storm evolution conditioned strictly on an initial observation (prompt) $\mathbf{x}_0(0)$. To obscure the future from the model, the forward process forms the noisy sample $\mathbf{x}_n$ by setting all non-initial frames to pure noise ($\mathbf{x}_n(1,T-1)=\boldsymbol{\epsilon}(1,T-1)$), while the initial frame receives standard diffusion noise,
$ \mathbf{x}_n(0)=\sqrt{\bar{\alpha}_n} \mathbf{x}_0(0) + \sqrt{1 - \bar{\alpha}_n} \boldsymbol{\epsilon}(0). $

During the reverse denoising process, we anchor the temporal generation by re-injecting the un-noised prompt $\mathbf{x}_0(0)$ at every step of the sampling sequence \citep{lugmayr2022repaint,ho2022video}. This process ensures that the generated storm remains tied to the initial prompt.

\section{Evaluations} \label{sec:evals}
\vspace{-0.3cm}

To assess the capabilities of GeoDES, our evaluation framework is designed to directly address the core limitations of current meteorological machine learning models. Specifically, we design our experiments to validate the following four claims: \vspace{-.3cm}
\begin{enumerate}[topsep=4pt,itemsep=2pt,leftmargin=*,parsep=0pt]
    \item \textbf{GeoDES outperforms prior works in storm realism.} Because the utility of synthetic weather data depends entirely on its physical plausibility, we verify that our architecture captures atmospheric dynamics more accurately than existing state-of-the-art baselines.
    \item \textbf{GeoDES' realistic storm synthesis capabilities are especially apparent for \textit{extreme} storms.} The primary motivation of this work is to augment datasets of rare, high-impact weather hazards. Therefore, we prove that the model does not fail on the most severe (and most difficult to synthesize) events.
    \item \textbf{GeoDES maintains reasonable compute requirements during training and inference.} A key limitation of global weather models is their immense computational cost. We demonstrate that our localized, storm-centered approach is accessible on standard hardware.
    \item \textbf{Ablations: Individual components are critical to GeoDES' success.} 
\end{enumerate}
\vspace{-0.25cm}

We detail our evaluation strategy, including metrics and baselines, in Section \ref{sec:evals-setup}, then validate our claims sequentially in Sections~\ref{sec:exp1} through \ref{sec:exp4}. Definitions, while provided throughout, are also listed in Appendix~\ref{app:glossary}.
\vspace{-0.8cm}

\subsection{Conditional Storm Synthesis: Evaluation Setup}\label{sec:evals-setup}
\vspace{-0.3cm}

\begin{wrapfigure}{r}{0.4\textwidth}
    \vspace{-0.8cm}
    \centering
    \includegraphics[width=\linewidth]{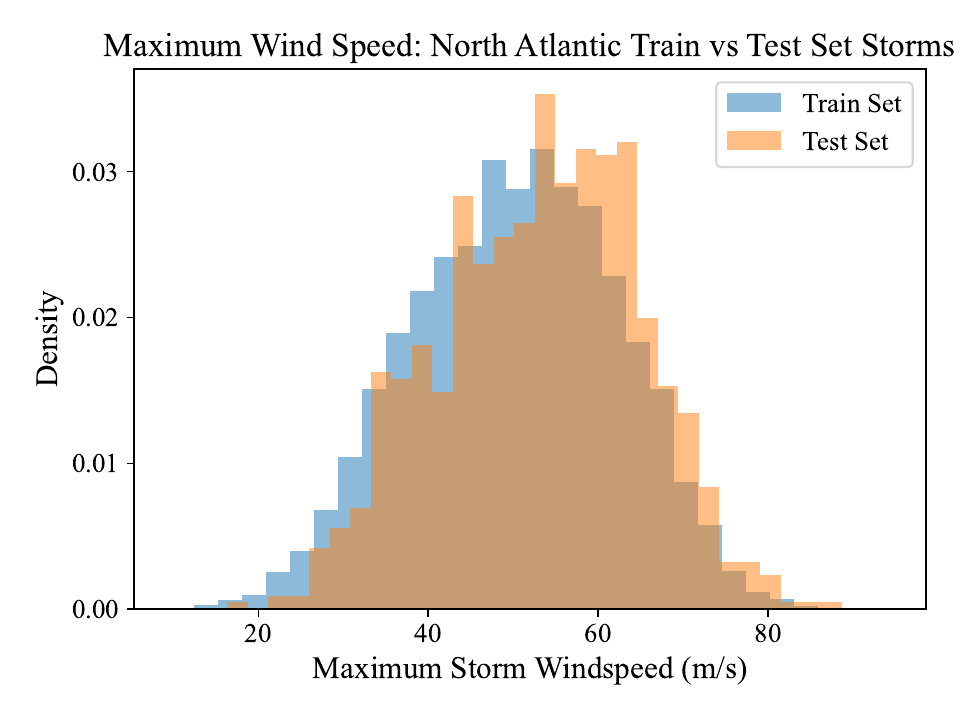}
    \vspace{-0.6cm}
    \caption{Normalized distribution of maximum 500hPa wind speeds for train vs test set. The test set exhibits a slightly higher intensity bias, reflecting recent trends.}
    \label{fig:train-v-test-histo}
    \vspace{-0.3cm}
\end{wrapfigure}
\textbf{Dataset:} We focus on \textit{extratropical cyclones}, which are mid-latitude low-pressure systems that are primary drivers of global atmospheric energy transfer \citep{martin2013mid}. Forming along surface temperature gradients (fronts) \citep{https://doi.org/10.1002/2016RG000519}, they are associated with severe winds and heavy precipitation \citep{hawcroft_how_2012, attinger_systematic_2021}, making them at times highly destructive weather events \citep{steiner2021regional}.

Global atmospheric data is sourced from the ERA5 reanalysis dataset \citep{hersbach2020era5}, with training data from January 1940 to August 2015 and test data from January 2016 through December 2024. This partition prevents co-occurrence leakage in the global model baselines, which may otherwise observe a test set storm during training if a separate train set storm happens to occur simultaneously. As a byproduct of shifting global climate, this split induces a slight distributional shift towards intense storms in the test set (Figure~\ref{fig:train-v-test-histo}). 

As our target use-case is study and evaluation by climate scientists, we focus on a lightweight selection of five climate variables useful to analyzing an extratropical cyclone: sea-level pressure (SLP), U- and V-components of wind at the 500hPa pressure-level, temperature at 925hPa and humidity at 500hPa. While global models are trained on the full training time span and permitted to view the full global data, cyclone-centered baselines are trained solely on cyclone-centered bounding boxes such as Figure~\ref{fig:compare}. At test-time, the outputs of cyclone-centered models are used directly, while cyclone-centered bounding boxes are first extracted from global models' outputs. More details on dataset processing and the creation of cyclone-centered bounding boxes are available in Appendix~\ref{app:preprocess-data}.

% \newpage
\textbf{Metrics:} We evaluate our architectures using a hybrid suite of metrics drawn from both deep learning and meteorology, broadly categorized into three areas:\vspace{-0.1cm}

Physical and Structural Coherence: \vspace{-0.15cm}
\begin{itemize}[topsep=0pt,itemsep=-0.01cm,leftmargin=*,partopsep=4pt, parsep=4pt]
    \item \textit{Peak Relative Vorticity Error:} The maximum rotational intensity of the cyclone via spatial gradients ($\zeta = \frac{dv}{dx} - \frac{du}{dy}$), identifying models that fail to generate tight, organized cyclones \citep{holton2012dynamic}.
    \item \textit{High-Frequency (HF) Spectral Ratio:} The ratio of integrated energy in the high-frequency domain of the Power Spectral Density between the prediction and ground truth \citep{skamarock2004evaluating, ravuri2021skilful}. This metric exposes models that output overly-smoothed, safe predictions, or inflate intensity scores by outputting unconstrained, high-frequency noise.  An ideal ratio is 1.0.
\end{itemize}

Extreme Intensity and Distribution: \vspace{-0.2cm}
\begin{itemize}[topsep=0pt,itemsep=-0.01cm,leftmargin=*,partopsep=4pt, parsep=4pt]
    \item \textit{Frequency Bias Index (FBI):} The ratio of predicted extreme wind pixels ($>20$ m/s) to actual extreme wind pixels \citep{wilks2011statistical}. An FBI of 1.0 indicates perfect extreme energy recovery, while $>1.0$ indicates generative hallucination (over-forecasting) and $<1.0$ indicates unphysical smoothing.
    \item \textit{Fractions Skill Score (FSS):} A neighborhood-based spatial verification metric (using a $3 \times 3$ grid cell window) that measures the spatial overlap accuracy specifically for the extreme ($>20$ m/s) wind distribution \citep{roberts2008scale}. FSS is formulated as 1 minus the ratio of the fractional Mean Squared Error (MSE) to a worst-case reference MSE, yielding a score strictly bounded between $0.0$ (no overlap) and $1.0$ (perfect overlap).
\end{itemize}
\vspace{-0.1cm}

Spatial and Navigational Accuracy: \vspace{-0.2cm}
\begin{itemize}[topsep=0pt,itemsep=-0.01cm,leftmargin=*,partopsep=4pt, parsep=4pt]
\item \textit{Synoptic Anomaly Correlation Coefficient (ACC):} Measures the spatial pattern correlation of the predicted fields \citep{wilks2011statistical, jolliffe2012forecast}. 
To avoid penalizing structurally sound storms that are offset from ground-truth examples, we apply a spatial Gaussian filter ($\sigma=1.5$) prior to correlation, allowing for a fair comparison against heavily smoothed autoregressive baselines.
\end{itemize}

We also provide average and per-variable root mean square error (RMSE), synoptic and non-synoptic ACC scores in Appendix~\ref{app:altmetrics}. Our primary metrics listed above were selected to give a fair evaluation of weather synthesis, rather than forecasting performance: Pixel-wise metrics (e.g., RMSE) can be misleading for weather synthesis tasks, penalizing realistic features that are slightly displaced compared to ground truth, through a phenomenon known as the ``double penalty'' problem \citep{roberts2008scale, ebert2008fuzzy}. 

\begin{table}[]
\caption{Comparison of the methods evaluated in Section~\ref{sec:evals}. The first row specifies the underlying architecture of each model, the second row lists whether the model operates directly on pixel space or uses an additional encoder model, while the third row states whether the model is a pretrained WFM.}
\vspace{-0.3cm}
\label{tab:baselines-summary} \small \centering
\renewcommand{\arraystretch}{0.75}
\begin{NiceTabular}{@{}r
>{\columncolor[HTML]{c3dcfe}}c ccccc@{}}
\toprule
                       & \textbf{GeoDES} & SVD       & CEF       & CoDiCast  & ClimaX      & Aurora      \\ \midrule
Architecture           & Diffusion       & Diffusion & Diffusion & Diffusion & Transformer & Transformer \\
Encoder vs Pixel Space & Pixel           & Pixel     & Pixel     & Encoder   & Pixel       & Encoder     \\
Pretrained Foundation Model       & No              & No        & No        & No        & Yes         & Yes   \\ \bottomrule    
\end{NiceTabular} 
\vspace{-0.8cm}
\end{table}

\textbf{Methods:} We evaluate GeoDES against a diverse suite of baselines, including statistical climatology, prior diffusion-based architectures and pretrained global Weather Foundation Models (WFMs). To ensure relevance for computationally-accessible, specialized storm analysis, these baselines were selected according to three criteria: (1) feasibility of execution on a single GPU; (2) the capacity to be trained and sampled using a sparse, five-variable atmospheric state rather than a full deep-column ERA5 initialization; and (3) the availability of a public codebase and, for WFMs, pretrained weights. We describe each baseline below. \vspace{-0.2cm}

\begin{itemize}[topsep=0pt,itemsep=0cm,leftmargin=*,partopsep=0pt, parsep=4pt]
    \item \textit{Climatology:} As a naive baseline, we simply average all  training datapoints together, forming a single composite storm climatology \citep{rasp2020weatherbench}. We predict this average storm for all test datapoints. % We use a climatology \citep{rasp2020weatherbench} calculated from the ERA5 record (1940–2024) as a baseline of minimal predictive skill. Using global six-hourly data, we compute monthly averages across all years and extract synthetic storm patches from the climatological month(s) corresponding to the ground-truth event. Including the test period creates advantaged ``oracle'' climatology, so that GeoDES’s performance is measured against a strictly competitive, rather than historically degraded, anchor.

    \item \textit{Stable Video Diffusion (SVD):} \cite{blattmann_stable_2023} represents an off-the-shelf diffusion baseline. We train a randomly initialized SVD model in pixel-space on linearly normalized data. Similar to GeoDES, time is represented as a native dimension within the model, enabling non-autoregressive sampling.
\end{itemize}

The remainder of our evaluation suite is composed of global forecasting models, selected due to the lack of modern patch-based baselines. To adapt these methods to our setting, we execute their training and inference scripts over global ERA5 data. For each storm, we extract the initial frame ($t=0$) from the spatial coordinates of the ground truth storm, then track the storm through the synthetic global forecast via minimum Sea Level Pressure (SLP). More details are in Appendix~\ref{app:preprocess-data}. \vspace{-0.2cm}

\begin{itemize}[topsep=0pt,itemsep=0cm,leftmargin=*,partopsep=0pt, parsep=4pt]
    \item \textit{Continuous Ensemble Forecasting (CEF):} \cite{andrae2025continuous} is a diffusion-based global forecasting ensemble operating on a 5.625$^{\circ}$ grid. Conditioned on an initial global state, CEF maps the forecast into a latent space and uses a continuous deterministic ODE solver to generate ensemble trajectories in parallel. %temporally consistent ensemble trajectories in parallel, bypassing the need for autoregressive steps for each ensemble member.

    \item \textit{CoDiCast:} \cite{shi2025codicast} is a diffusion-based global forecasting model designed to operate on a 5.625$^{\circ}$ grid. Conditioned on an initial state encoded via a pretrained deterministic autoencoder, CoDiCast autoregressively predicts each subsequent atmospheric timestep.

    \item \textit{ClimaX WFM (Public and Fine-tuned Variants):} \cite{nguyen2023climax} is a transformer-based global WFM. Because pretrained ClimaX does not offer SLP, we instead derive SLP from ClimaX geopotential and 2-meter temperature predictions---see Appendix~\ref{app:preprocess-data}. We use the highest-resolution checkpoint (1.40625$^{\circ}$). ClimaX's pretraining set spans January 1940-December 2015, which aligns with our train/test split. We include two variants of ClimaX in our evaluations: \textit{ClimaX FT}, which we fine-tune on our five-variable subset of ERA5, and the publicly-available \textit{ClimaX}, without additional fine-tuning.
    
    \item \textit{Aurora WFM:} \cite{bodnar2024aurora} is a transformer-based global WFM with an additional encoder. We use the 0.25$^{\circ}$ checkpoint. We note that publicly available Aurora checkpoints are fine-tuned on 2016–2021 data. Consequently, evaluation on our 2016–2024 test set contains temporal data leakage. Even with this artificial advantage, we demonstrate in Sections~\ref{sec:exp1} and \ref{sec:exp2} that off-the-shelf global WFMs fail to resolve the fine-grained rotational dynamics of extratropical cyclones.
\end{itemize}

\subsection{Storm Modeling Capabilities}\label{sec:exp1}

We begin by validating our first claim.\\
\textbf{Claim 1:} GeoDES outperforms prior modeling approaches in storm realism.

% \newpage

\begin{table}[b]
\caption{Comparison of generative storm realism for North Atlantic extratropical cyclones. GeoDES successfully maintains physical constraints and extreme energy distributions, avoiding both the high-frequency hallucination of SVD and the spatial smoothing of global Weather Foundation Models. In the far left column, ($\downarrow$) and ($\uparrow$) respectively indicate lower or higher scores are better, while ($\cdot 1$) indicates that scores closer to $1$ are better. Best results are bolded.} \vspace{-.3cm}
\label{tab:exp-1}
\resizebox{\textwidth}{!}{%
\begin{NiceTabular}{@{}r
>{\columncolor[HTML]{c3dcfe}}c cccccc
>{\columncolor[HTML]{EFEFEF}}c }
\toprule
                                    & \textit{\textbf{GeoDES}}    & SVD                         & CoDiCast               & CEF            & ClimaX FT      & ClimaX  & Aurora  &  \textit{Climatology} \\ \midrule
Peak Vorticity Error ($\downarrow$) & \textbf{$\bf{5.04\pm0.47}$} & $48.40\pm0.02$              & $10.52\pm1.72$ & $11.04\pm0.77$ & $10.68\pm1.01$ & $12.32$ & $13.80$ & $\it 7.86$      \\
HF Spectral Ratio ($\cdot 1$)       & \textbf{$\bf{0.95\pm0.17}$} & $52.52\pm0.00$              & $3.65\pm1.09$  & $0.59\pm0.02$  & $0.41\pm0.02$  & $0.02$  & $0.00$  & $\it 0.29$      \\
Frequency Bias Index ($\cdot 1$)    & \textbf{$\bf{1.02\pm0.17}$} & $1.10\pm0.00$               & $2.26\pm0.09$  & $0.60\pm0.05$  & $1.86\pm0.03$  & $1.04$  & $0.00$  & $\it 0.75$      \\
Fractions Skill Score ($\uparrow$)  & \textbf{$\bf{0.76\pm0.04}$} & \textbf{$\bf{0.76\pm0.00}$} & $0.65\pm0.01$  & $0.61\pm0.01$  & $0.72\pm0.01$  & $0.00$  & $0.00$  & $\it 0.71$      \\
Synoptic ACC ($\uparrow$)           & \textbf{$\bf{0.39\pm0.03}$} & $0.15\pm0.00$               & $0.06\pm0.01$  & $0.19\pm0.02$  & $0.28\pm0.00$  & $0.09$  & $0.10$  & $\it 0.64$      \\
Synoptic ACC T-Final ($\uparrow$)   & \textbf{$\bf{0.44\pm0.03}$} & $0.14\pm0.00$               & $0.06\pm0.01$  & $0.18\pm0.02$  & $0.24\pm0.00$  & $0.05$  & $0.11$  & $\it 0.64$      \\ \bottomrule
\end{NiceTabular}%
} 
% \vspace{-.5cm}
\end{table}

\begin{figure}
    \centering
    \includegraphics[width=0.95\textwidth]{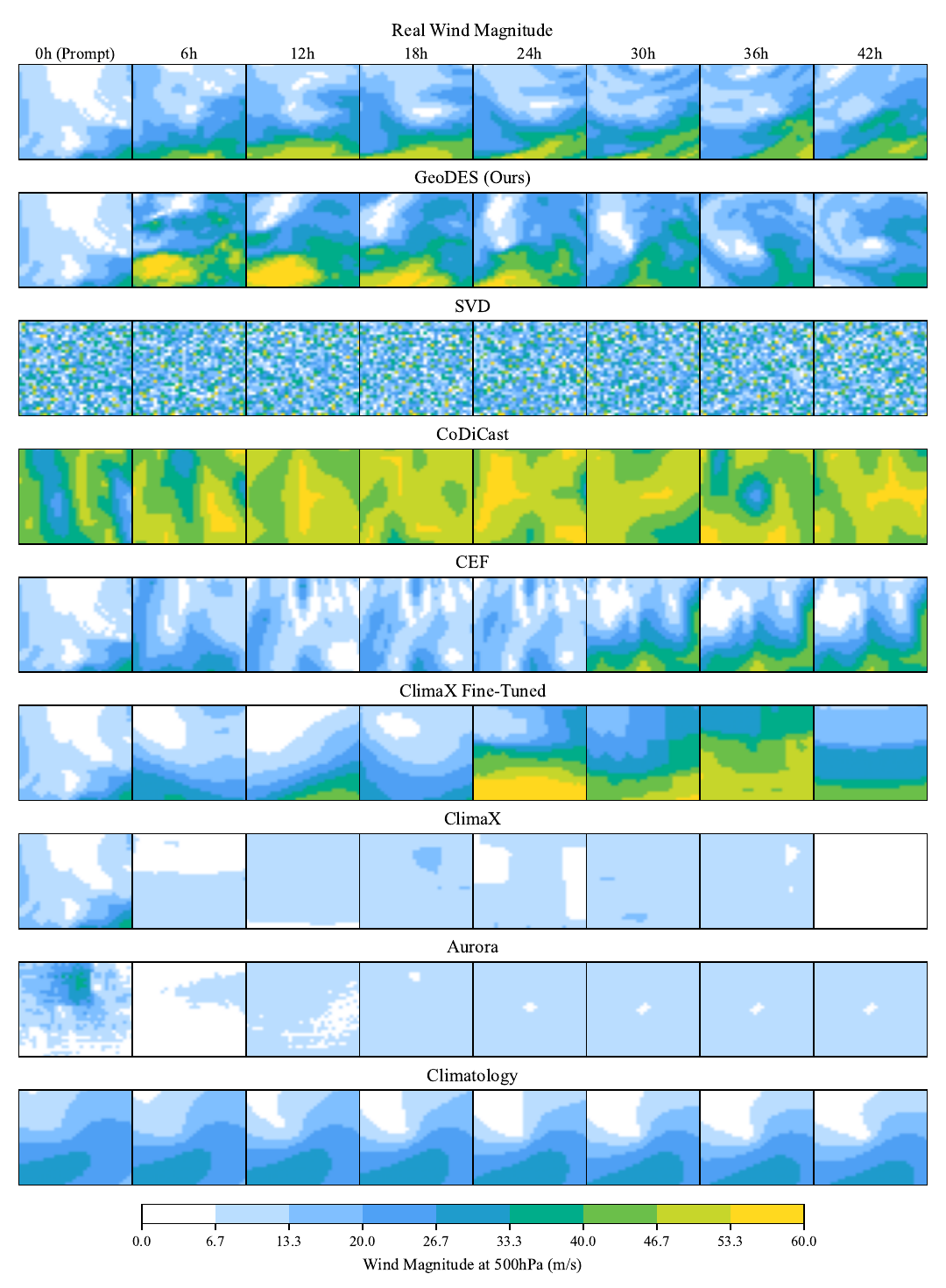}
    \vspace{-0.5cm}
    \caption{Wind magnitude of a real extratropical cyclone in the North Atlantic ($57.60^\circ$ N, $26.87^\circ$ W at 0h) on October 3-5, 2018, and corresponding output from each synthesis method. GeoDES creates a realistic, novel storm, while other methods generate blurry (ClimaX, Aurora) or noisy (SVD, CoDiCast) outputs.}
    \label{fig:compare}
\end{figure}

\textbf{Setup:} Table~\ref{tab:exp-1} includes results for our primary metrics across all evaluated methods on the North Atlantic basin test set ($n=893$). Hyperparameter selection information for GeoDES and baselines is available in Appendix~\ref{app:hyperparams}. Apart from the inference-only methods (ClimaX, Aurora and climatology), evaluations are averaged across three training/sampling runs. As an example of each model's outputs, Figure~\ref{fig:compare} visualizes synthesized wind magnitude for the same real storm initial conditions. In the figure, each model has been prompted with the 0h conditions of the real storm sample in the top row. Additional metrics are presented in Appendix~\ref{app:altmetrics} and Table~\ref{tab:exp1-supp}.  In Appendix~\ref{app:wholeworld}, we also evaluate GeoDES on additional basins.

\textbf{Results:} \textbf{\textit{GeoDES outperforms probabilistic methods across evaluation metrics.}} In particular, GeoDES is the only model capable of recovering the kinetic energy distribution of ground-truth storms, achieving a near-perfect Frequency Bias Index of $1.02$ and a High-Frequency (HF) Spectral Ratio of $0.95$. 

Meanwhile, baselines present one of two contrasting failure modes. Off-the-shelf SVD collapses into overly-noisy predictions, injecting over $50$ times the natural high-frequency energy into its synthetic storms (HF Spectral Ratio of $52.52$) while failing to represent a cyclone storm structure, resulting in a Peak Vorticity Error of $48.40$. This collapse is a consequence of its approaches to data normalization and temporal processing, which are better-adapted to RGB image generation than to atmospheric data. CoDiCast presents a similar noisy failure mode, while the other global weather models of CEF, ClimaX FT, ClimaX and Aurora fall into a pattern of low-energy, blurry predictions. This pattern is generally characterized by low HF Spectral Ratio and Synoptic ACC, as well as moderately high Peak Vorticity Error.

It is worth noting the deceptively high Synoptic ACC scores achieved by the climatology baseline. As a smooth average of all training storms, it captures the broad, low-frequency information of a cyclone while lacking sharp, high-intensity gradients. Consequently, it avoids the severe spatial  penalty that dynamic models face when predicting fine-grained features that are even slightly displaced. While this smoothing allows the composite to game spatial correlation metrics, its low HF Spectral Ratio and poor Frequency Bias Index confirm it lacks the physical realism and kinetic energy required for actual storm synthesis.

Successes and failure modes of each method are also demonstrated by their outputs in Figure~\ref{fig:compare}. While GeoDES (second row) synthesizes a realistic, detailed and dynamic extratropical cyclone, the fine-tuned global models (CoDiCast, CEF and ClimaX FT) present odd distortions indicative of a failure to model high-frequency storm attributes. Worse, the zero-shot models (ClimaX and Aurora) present very blurry predictions that somewhat resemble the climatology baseline. Due to reliance on encoders, rather than operating in pixel-space, CoDiCast and Aurora fail to reproduce the 0h prompt conditions altogether. SVD, our only cyclone-centered baseline, presents the most significant collapse: while the top third of each frame perhaps represents the transition from calm seas to a storm front, the bottom half of each frame contains significant amounts of static noise indicative of collapse and inability to recreate accurate storm features.

These findings contextualize the baselines' original design objectives. \textit{Existing ML weather models primarily assess their performance via grid-wise error metrics (e.g., RMSE) rather than the localized generative fidelity of specific severe weather phenomena like extratropical cyclones}. Because optimizing for pixel-wise error incentivizes spatial mean-seeking, these models are predisposed to smoothing.

% \newpage
\vspace{-0.3cm} 
\subsection{Severe Storm Representation}\label{sec:exp2}

\begin{table}[b]
\vspace{-0.3cm} 
\caption{Comparison of generative storm realism for the subset of most extreme North Atlantic extratropical cyclones (defined as top $10\%$ of storm-lifetime wind magnitude). GeoDES maintains its strong synthesis performance demonstrated in Table~\ref{tab:exp-1} in this more challenging setting. In the far left column, ($\downarrow$) and ($\uparrow$) respectively indicate lower or higher scores are better, while ($\cdot 1$) indicates that scores closer to $1$ are better. Best results are bolded.} \vspace{-0.3cm}
\label{tab:exp-2}
\resizebox{\textwidth}{!}{%
\begin{NiceTabular}{@{}r
>{\columncolor[HTML]{c3dcfe}}c cccccc
>{\columncolor[HTML]{EFEFEF}}c }
\toprule
                                    & \textit{\textbf{GeoDES}}    & SVD                         & CoDiCast      & CEF            & ClimaX FT      & ClimaX  & Aurora  &  \textit{Climatology} \\ \midrule
Peak Vorticity Error ($\downarrow$) & \textbf{$\bf{4.47\pm1.04}$} & $47.08\pm0.02$              & $6.90\pm2.28$ & $11.80\pm0.95$ & $11.14\pm1.18$ & $12.96$ & $14.37$ & $\it{8.28}$      \\
HF Spectral Ratio ($\cdot 1$)       & \textbf{$\bf{0.71\pm0.14}$} & $36.00\pm0.00$              & $2.02\pm0.94$ & $0.36\pm0.02$  & $0.27\pm0.01$  & $0.01$  & $0.00$  & $\it{0.19}$      \\
Frequency Bias Index ($\cdot 1$)    & $0.83\pm0.13$               & \textbf{$\bf{0.88\pm0.00}$} & $1.79\pm0.08$ & $0.42\pm0.06$  & $1.55\pm0.08$  & $0.00$  & $0.00$  & $\it{0.59}$      \\
Fractions Skill Score ($\uparrow$)  & \textbf{$\bf{0.82\pm0.07}$} & $0.78\pm0.00$               & $0.75\pm0.01$ & $0.53\pm0.04$  & $0.81\pm0.01$  & $0.00$  & $0.00$  & $\it{0.79}$      \\
Synoptic ACC ($\uparrow$)           & \textbf{$\bf{0.45\pm0.05}$} & $0.16\pm0.00$               & $0.06\pm0.01$ & $0.18\pm0.02$  & $0.34\pm0.00$  & $0.12$  & $0.13$  & $\it{0.70}$      \\
Synoptic ACC T-Final ($\uparrow$)   & \textbf{$\bf{0.50\pm0.06}$} & $0.17\pm0.00$               & $0.04\pm0.02$ & $0.18\pm0.01$  & $0.28\pm0.01$  & $0.05$  & $0.14$  & $\it{0.70}$      \\  \bottomrule
\end{NiceTabular}%
} 
% \vspace{-0.4cm}
\end{table}

\textbf{Claim 2:} GeoDES' realistic storm synthesis capabilities are especially apparent for \textit{extreme} storms.

\textbf{Setup:} We follow a similar experimental setup to Section~\ref{sec:exp1}, evaluating strictly on the 90th percentile of most extreme storms within the North Atlantic basin test set (n=89), defined by storm-lifetime maximum wind magnitude \citep{doi:10.1126/science.aav9527}. Given the relative rarity of extreme training examples, compounded by the out-of-distribution severity bias of the test set (see Figure~\ref{fig:train-v-test-histo}), this subset represents a challenging synthesis task. Results are presented in Table~\ref{tab:exp-2} with supplemental metrics in Appendix~\ref{app:altmetrics} and Table~\ref{tab:exp2-supp}.

\textbf{Results:} Under the extreme storm setting, the performance gap between GeoDES and the baseline models  widens regarding physical realism and the representation of extreme-energy behavior. 
While ClimaX FT achieves a highly competitive Fractions Skill Score (0.81 compared to GeoDES's 0.82) and the climatology baseline continues to display misleading ACC scores (refer to Section~\ref{sec:exp1}), 
most methods struggle to accurately represent the kinetic energy and structural integrity of severe storms. For example, the global WFMs (ClimaX and Aurora) experience catastrophic spatial smoothing. Both models produce FBI scores of 0.00 and HF Spectral Ratios near 0.00, indicating that their predictions are overly regularized by the climatological mean, preventing them from synthesizing extreme weather. Conversely, off-the-shelf SVD attempts to match the ground-truth high-intensity of these storms, but completely fails to constrain the physics; it generates 36 times the natural high-frequency atmospheric energy (HF Spectral Ratio of 36.00) and suffers massive rotational deformation (Peak Vorticity Error of 47.08), indicating severe generative hallucination rather than coherent storm synthesis.

In contrast, \textbf{GeoDES generates realistic extreme storms.} GeoDES matches or slightly exceeds its full-dataset performance on key structural metrics. Despite the severity of this data subset, GeoDES correctly preserves the rotational core (Peak Vorticity Error of $4.47\pm 1.04$) and safely recovers the vast majority of the extreme wind distribution (Frequency Bias Index of $0.83\pm 0.13$). This performance demonstrates that the GeoDES architecture correctly parameterizes severe weather without reverting to the mean or hallucinating unphysical noise.

\subsection{Computational Requirements}\label{sec:exp3}
\vspace{-0.4cm}

\textbf{Claim 3:} GeoDES maintains reasonable compute requirements during training and inference.

\textbf{Setup:} For each model listed in Section~\ref{sec:evals-setup}, Table~\ref{tab:compute} reports the model parameter count, sampling paradigm (per-storm versus continuously over the 2016--2024 test span) and the resulting number of inference steps, as well as the FLOPs, latency, VRAM and energy required to produce one $42$-hour sample. Model parameter count and FLOPs are both calculated using the Python \texttt{fvcore} library implementations for all models except the TensorFlow-implemented CoDiCast, for which we calculate these metrics from scratch. All measurements are performed on the same machine with an NVIDIA RTX A6000 GPU.

Due to the prohibitive System RAM requirements of tracing Aurora's execution graphs for $0.25^\circ$ global inputs (>1M spatial tokens), exact FLOP counts for Aurora could not be profiled natively. We instead estimate FLOPs as a function of parameter count and sequence size according to transformer scaling laws \cite{kaplan2020scalinglawsneurallanguage}. Refer to Appendix~\ref{app:compute} for additional details.

As a signal of train-time compute requirements, we also measure the 2D spatial model used to train GeoDES. To sample the same number of frames as the 3D model (i.e., $T=8$), the 2D model uses $9.4$ TFLOPs.

\begin{table}[] \centering
\caption{Test-time computational requirements for each evaluated model, including model size, sampling paradigm, inference steps, FLOPs, latency, VRAM and energy to produce one $42$-hour sample. While GeoDES contains more parameters than other models, all other metrics such as latency and energy usage are competitive with other models. *: FLOPs could not be empirically measured for Aurora due to system RAM limitations; please see estimate derivation in Appendix~\ref{app:compute}.} 
\label{tab:compute} \small \vspace{-0.3cm}
\renewcommand{\arraystretch}{0.85}
\begin{NiceTabular}{@{}r
>{\columncolor[HTML]{c3dcfe}}c ccccc}
\toprule
            & \textit{\textbf{GeoDES}} & SVD            & CoDiCast & CEF    & ClimaX & Aurora       \\ \midrule
Params (M)       & $1802$                   & $1525$ & $63$       & $56$       & $111$      & $1256$         \\
Paradigm         & Storm                    & Storm  & Continuous & Continuous & Continuous & Continuous     \\
Inference Steps & $893$                      & $893$    & $13145$      & $13145$      & $13145$      & $13145$          \\
FLOPs (T)        & $123.7$                  & $57.6$ & $93.5$     & $34.6$     & $2.2$      & $\sim 4557$* \\
Latency (s)      & $17.3$                   & $7.7$  & $1029.2$   & $4.1$      & $0.9$      & $47.6$         \\
VRAM (GB)        & $7.1$                    & $6.0$  & $1.1$      & $0.3$      & $1.6$      & $13.4$         \\
Energy (Wh)      & $1.44$                   & $0.64$ & $85.77$    & $0.34$     & $0.07$     & $3.97$   \\ \bottomrule
\end{NiceTabular} \vspace{-0.8cm}
\end{table}

\textbf{Results:} \textit{At test time:} GeoDES demonstrates practical inference requirements. While GeoDES has the largest parameter footprint of the evaluated models ($1.8$B), the storm-centered approach restricts its computation strictly to the spatial bounding box of the storm. This design choice bypasses the large compute explosion characteristic of high-resolution global Weather Foundation Models (WFMs). For instance, while Aurora contains fewer parameters ($1.25$B) than GeoDES, processing its high-resolution global grid requires over $36$ times as many FLOPs and nearly double the VRAM ($13.4$ GB).

GeoDES also avoids computational traps of autoregressive architectures. Despite CoDiCast's smaller parameter count ($63$M), its autoregressive sequence generation results in severe per-sample latency ($1029.2$s) and energy consumption ($85.77$ Wh), whereas GeoDES uses a fraction of the time ($17.3$s) and energy ($1.44$ Wh). Continuous evaluation models (CoDiCast, CEF, ClimaX, Aurora) further compound this cost over the 2016--2024 period, performing over $13,000$ inference steps. Consequently, heavier architectures like CoDiCast cumulatively exceed $1$ MWh, heavily outpacing the storm-centric paradigm of GeoDES that is only prompted once for each of the $893$ storms in the test set.

Ultimately, GeoDES maintains highly accessible compute requirements that enable its use on a single standard commercial GPU. While heavily smoothed global models like ClimaX offer lower absolute latency, GeoDES strikes a more reliable balance between generative capability (demonstrated in Sections~\ref{sec:exp1} and \ref{sec:exp2}) and practical computational cost for storm-centric tasks.

\textit{At train time:} GeoDES' temporal inflation scheme results in significant time and computational resource savings compared to training the 3D spatiotemporal model from scratch. Because half of the GeoDES training epochs are allocated to the 2D stage, we reduce the computational bottleneck associated with spatiotemporal diffusion. In particular, because the 2D spatial generation phase requires only $9.4$ TFLOPs per sample, compared to the $123.7$ TFLOPs at 3D temporal phase, GeoDES achieves its final performance with only $53.8\%$ of the total theoretical compute that would have been required to train the 3D architecture from scratch.

\subsection{Ablations}\label{sec:exp4}

\begin{table}[]
\caption{Ablation study of the GeoDES architecture evaluated on the North Atlantic test set. We evaluate models trained without signal-to-noise ratio weighting (No SNR), without the correlated noise schedule (Non-Stoch.), using standard linear scaling rather than the variable-specific hybrid normalizer (Linear Norm.) and training the spatiotemporal architecture entirely from scratch (3D W/Out 2D). Results demonstrate that the entire GeoDES pipeline contributes to physical realism, with the 2D-to-3D temporal inflation phase proving especially critical for overall model stability. Best results are bolded.}  \vspace{-0.3cm}
\label{tab:ablations}
\resizebox{\textwidth}{!}{%
\begin{NiceTabular}{@{}r
>{\columncolor[HTML]{c3dcfe}}c cccc}
\toprule
                                    & \textit{\textbf{GeoDES}}    & No SNR                 & Non-Stoch.    & Linear Norm.                & 3D W/Out 2D   \\ \midrule
Peak Vorticity Error ($\downarrow$) & \textbf{$5.04\pm0.47$}      & $5.21\pm0.36$          & $5.43\pm0.28$ & \textbf{$\bf{4.96\pm0.05}$} & $5.07\pm0.04$ \\
HF Spectral Ratio ($\cdot 1$)          & \textbf{$\bf{0.95\pm0.17}$} & $0.71\pm0.19$          & $0.76\pm0.08$ & $0.79\pm0.11$               & $1.09\pm0.35$ \\
Frequency Bias Index ($\cdot 1$)          & \textbf{$\bf{1.02\pm0.17}$} & \textbf{$0.93\pm0.08$} & $0.95\pm0.04$ & $1.06\pm0.12$               & $1.10\pm0.19$ \\
Fractions Skill Score ($\uparrow$)   & \textbf{$\bf{0.76\pm0.04}$} & $0.73\pm0.05$          & $0.72\pm0.02$ & \textbf{$\bf{0.76\pm0.00}$} & $0.62\pm0.08$ \\
Synoptic ACC ($\uparrow$)           & \textbf{$\bf{0.39\pm0.03}$} & $0.36\pm0.04$          & $0.37\pm0.00$ & $0.37\pm0.01$               & $0.33\pm0.01$ \\
Synoptic ACC T-Final ($\uparrow$)   & \textbf{$\bf{0.44\pm0.03}$} & $0.41\pm0.04$          & $0.39\pm0.01$ & $0.42\pm0.00$               & $0.38\pm0.02$ \\ \bottomrule
\end{NiceTabular}%  \vspace{-0.4cm}
}
\end{table}

\textbf{Claim 4:} Ablation analysis demonstrates the effectiveness of GeoDES as a whole.

\textbf{Setup:} We test four ablation variants of GeoDES: 
\begin{itemize}[nosep]
    \item \textit{No SNR:} Signal-to-noise weighting is disabled.
    \item \textit{Non-Stoch.:} The model is sampled deterministically without correlated noise.
    \item \textit{Linear Norm.:} All atmospheric channels are normalized linearly.
    \item \textit{3D W/Out 2D:} A randomly initialized spatiotemporal GeoDES model is trained entirely from scratch for the total number of training epochs.
\end{itemize}
Primary evaluation metrics are in Table~\ref{tab:ablations}; supplementary metrics are in Appendix~\ref{app:altmetrics} and Table~\ref{tab:exp4-supp}.

\textbf{Results:} The ablation study confirms that all components of the GeoDES pipeline are necessary to achieve high-quality storm synthesis.

\textit{Temporal Inflation (3D W/Out 2D):} Training the 3D spatiotemporal model from scratch yields significant performance collapse, particularly in spatial overlap and navigational accuracy. Without the spatial representation learned during 2D pretraining, Fractions Skill Score drops sharply from $0.76$ to $0.62$ and Synoptic ACC falls to $0.33$. This confirms that the two-stage training approach is not merely a computational optimization (as established in Section~\ref{sec:exp3}), but a critical requirement for spatial stability.

\textit{Hybrid Normalization (Linear Norm.):} While relying purely on linear normalization yields a competitive Peak Vorticity Error ($4.96$), the model fails to accurately preserve the energy distribution of extreme wind, demonstrated by its degraded HF Spectral Ratio ($0.79$ down from $0.95$). This demonstrates that applying log-scaling to long-tailed variables helps to prevent dynamic range collapse.

\textit{Noise and Sampling Mechanisms (No SNR \& Non-Stoch.):} Removing signal-to-noise ratio weighting or using deterministic sampling impairs the model's ability to represent fine-grained, chaotic dynamics of weather systems. Both ablations result in higher Peak Vorticity Errors ($>5.20$) and drops in high-frequency energy recovery (HF Spectral Ratios falling from $0.95$ to $0.71$ and $0.76$, respectively). This proves that properly weighted stochasticity is required to maintain realistic atmospheric turbulence.

\section{Related Work} \label{sec:relatedwork}

GeoDES is most closely related to the areas of synoptic storms research, weather modeling and diffusion architectures. We now provide more background on each of these areas.

\textbf{Synoptic storms research:} Synoptic storms are large-scale weather systems that span approximately $1000$km or more \citep{holton2012dynamic}. Well-known examples include extratropical cyclones, large hurricanes and atmospheric rivers. Decades of meteorology research has improved the understanding of how these storms occur and evolve, leading to improved forecasting capabilities. However, notable knowledge deficits remain, including the understanding of dangerous phenomena such as extratropical cyclone bombogenesis \citep{steiner2021regional} and the impact of climate change on synoptic storms \citep{gertler2019changing}. An effective paradigm for the study of synoptic storm behavior is the storm-centered approach, which involves the analysis of bounding boxes centered on the storm(s) of interest \citep{booth_extratropical_2018}. 

\textbf{Weather modeling:} Significant improvements over recent decades have resulted in a diversity of weather models. While traditional numerical weather models remain trusted for operational meteorology, machine learning has been increasingly applied towards the development of novel weather modeling approaches. Much research interest has been devoted to the development of weather foundation models (WFMs), with top-performing architectures including ClimaX \citep{nguyen2023climax}, Prithvi WxC \citep{schmude2024prithvi} and GraphCast \citep{lam2023learning}. Historically, ML-based weather models have been strictly divided into global or regional scopes \citep{shi2025deeplearningfoundationmodels}. Standalone regional models, such as \cite{andrychowicz2023deep} or \cite{mardani2024residual}, are generally dependent on land-based data and struggle to track the full lifecycle of a storm across large oceanic regions. Conversely, as we show in Section~\ref{sec:exp4}, deterministic global WFMs tend to smooth predictions of fine-grained storm details. While recent probabilistic frameworks, such as GenCast \citep{price2023gencast} and NeuralGCM \citep{kochkov2024neural}, mitigate this blurring through generative diffusion, they often impose prohibitive computational overheads for continuous high-resolution synthesis. Relatedly, deep learning has been applied to targeted aspects of extratropical cyclone analysis, such as estimating boundary-layer wind risks \citep{snaiki2022knowledge}. However, the efficient modeling of the complete spatiotemporal evolution of these synoptic storms has remained relatively under-explored.

\textbf{Diffusion architectures:} Diffusion modeling \citep{ho2020denoising} is a popular generative deep learning architecture often applied for image and video synthesis \citep{ho2022video, blattmann_stable_2023}; see Section~\ref{app:dm-background} for details. Diffusion models applied to weather modeling have focused on the problems of global forecasting, for instance CoDiCast \citep{shi2025codicast} and Continuous Ensemble Forecasting (CEF) \citep{andrae2025continuous}, and regional forecasting \citep{mardani2024residual}. Relatedly, diffusion models have been successfully applied to high-resolution precipitation nowcasting \citep{gao2023prediff} and the scalable generation of probabilistic forecast ensembles \citep{li2023seeds}.

\section{Limitations and Future Work}
\label{sec:futurework}

GeoDES demonstrates strong generative fidelity for large-scale storm structures at the 500hPa pressure level (representing the mid-level atmosphere), a standard and useful domain for analyzing storm evolution. However, future work should expand this framework to include precipitation and surface-level phenomena. Precipitation is a notorious challenge due to its sparse, discontinuous manner. Synthesizing 10-meter wind fields introduces additional complexities, such as boundary layer dynamics and topographical friction. Incorporating these extreme surface distributions is an important step for applying synthetic storms directly to infrastructural stress testing and hazard mitigation. Furthermore, transitioning GeoDES into a fully operational tool will require pairing the model with an intuitive inference pipeline, allowing meteorologists to easily condition, generate and interpret synthetic storms within standard climate science workflows.

\section{Conclusion} \label{sec:conclusion}
We present Geospatial Diffusion-based Evolution Synthesis (GeoDES), a model for the synthesis of large-scale storm structures. We demonstrate that GeoDES' storm-centered approach, 2D-to-3D temporal inflation strategy and weather-adapted data normalization enable realistic extratropical cyclone synthesis, useful to data augmentation and validation of downstream forecasting models.

\subsubsection*{Broader Impact Statement}
While we foresee few negative repercussions of GeoDES in particular, we note that all empirical deep learning research requires the consumption of energy which, in aggregate, negatively impacts the environment. We have taken efforts to be transparent about compute resources through Section~\ref{sec:exp3}.

\subsubsection*{Acknowledgments}
We thank the NASA Climate Change Research Initiative (CCRI), the National Science Foundation (NSF CCF-2106707), the Defense Advanced Research Projects Agency (DARPA) Young Faculty Award and the Wisconsin Alumni Research Foundation (WARF) for funding this work. We are also grateful to the University of Wisconsin-Madison Center for High Throughput Computing (CHTC) for providing GPU resources and to Kellie O'Grady for feedback on an early stage of this work.

\bibliography{main,zotero}

@article{ebert2008fuzzy,
  title={Fuzzy verification of high-resolution gridded forecasts: a review and proposed framework},
  author={Ebert, Elizabeth E},
  journal={Meteorological Applications},
  volume={15},
  number={1},
  pages={51--64},
  year={2008},
  publisher={Wiley Online Library}
}

@article{roberts2008scale,
  title={Scale-selective verification of rainfall accumulations from high-resolution forecasts of convective events},
  author={Roberts, Nigel M and Lean, HW},
  journal={Monthly Weather Review},
  volume={136},
  number={1},
  pages={78--97},
  year={2008},
  publisher={American Meteorological Society}
}

@book{wilks2011statistical,
  title={Statistical methods in the atmospheric sciences},
  author={Wilks, Daniel S},
  volume={100},
  year={2011},
  publisher={Academic press}
}

@book{jolliffe2012forecast,
  title={Forecast verification: a practitioner's guide in atmospheric science},
  author={Jolliffe, Ian T and Stephenson, David B},
  year={2011},
  publisher={John Wiley \& Sons}
}

@book{holton2012dynamic,
  title={An introduction to dynamic meteorology},
  author={Holton, James R and Hakim, Gregory J},
  volume={88},
  year={2012},
  publisher={Academic press}
}

@article{skamarock2004evaluating,
  title={Evaluating mesoscale NWP models using kinetic energy spectra},
  author={Skamarock, William C},
  journal={Monthly weather review},
  volume={132},
  number={12},
  pages={3019--3032},
  year={2004},
  publisher={American Meteorological Society}
}

@article{ravuri2021skilful,
  title={Skilful precipitation nowcasting using deep generative models of radar},
  author={Ravuri, Suman and Lenc, Karel and Willson, Matthew and Kangin, Dmitry and Lam, Remi and Mirowski, Piotr and Fitzsimons, Megan and Athanassiadou, Maria and Kashem, Sheleem and Madge, Simon and others},
  journal={Nature},
  volume={597},
  number={7878},
  pages={672--677},
  year={2021},
  publisher={Nature Publishing Group UK London}
}

@misc{hagemann2024efficientparallelizationlayoutslargescale,
      title={Efficient Parallelization Layouts for Large-Scale Distributed Model Training}, 
      author={Johannes Hagemann and Samuel Weinbach and Konstantin Dobler and Maximilian Schall and Gerard de Melo},
      year={2024},
      eprint={2311.05610},
      archivePrefix={arXiv},
      primaryClass={cs.LG},
      url={https://arxiv.org/abs/2311.05610}, 
}

@article{pope2023efficiently,
  title={Efficiently scaling transformer inference},
  author={Pope, Reiner and Douglas, Sholto and Chowdhery, Aakanksha and Devlin, Jacob and Bradbury, James and Heek, Jonathan and Xiao, Kefan and Agrawal, Shivani and Dean, Jeff},
  journal={Proceedings of machine learning and systems},
  volume={5},
  pages={606--624},
  year={2023}
}

@misc{chowdhery2022palmscalinglanguagemodeling,
      title={PaLM: Scaling Language Modeling with Pathways}, 
      author={Aakanksha Chowdhery and Sharan Narang and Jacob Devlin and Maarten Bosma and Gaurav Mishra and Adam Roberts and Paul Barham and Hyung Won Chung and Charles Sutton and Sebastian Gehrmann and Parker Schuh and Kensen Shi and Sasha Tsvyashchenko and Joshua Maynez and Abhishek Rao and Parker Barnes and Yi Tay and Noam Shazeer and Vinodkumar Prabhakaran and Emily Reif and Nan Du and Ben Hutchinson and Reiner Pope and James Bradbury and Jacob Austin and Michael Isard and Guy Gur-Ari and Pengcheng Yin and Toju Duke and Anselm Levskaya and Sanjay Ghemawat and Sunipa Dev and Henryk Michalewski and Xavier Garcia and Vedant Misra and Kevin Robinson and Liam Fedus and Denny Zhou and Daphne Ippolito and David Luan and Hyeontaek Lim and Barret Zoph and Alexander Spiridonov and Ryan Sepassi and David Dohan and Shivani Agrawal and Mark Omernick and Andrew M. Dai and Thanumalayan Sankaranarayana Pillai and Marie Pellat and Aitor Lewkowycz and Erica Moreira and Rewon Child and Oleksandr Polozov and Katherine Lee and Zongwei Zhou and Xuezhi Wang and Brennan Saeta and Mark Diaz and Orhan Firat and Michele Catasta and Jason Wei and Kathy Meier-Hellstern and Douglas Eck and Jeff Dean and Slav Petrov and Noah Fiedel},
      year={2022},
      eprint={2204.02311},
      archivePrefix={arXiv},
      primaryClass={cs.CL},
      url={https://arxiv.org/abs/2204.02311}, 
}

@misc{kaplan2020scalinglawsneurallanguage,
      title={Scaling Laws for Neural Language Models}, 
      author={Jared Kaplan and Sam McCandlish and Tom Henighan and Tom B. Brown and Benjamin Chess and Rewon Child and Scott Gray and Alec Radford and Jeffrey Wu and Dario Amodei},
      year={2020},
      eprint={2001.08361},
      archivePrefix={arXiv},
      primaryClass={cs.LG},
      url={https://arxiv.org/abs/2001.08361}, 
}

@article{cc1,
  title={A cyclone-centered perspective on the drivers of asymmetric patterns in the atmosphere and sea ice during Arctic cyclones},
  author={Clancy, Robin and Bitz, Cecilia M and Blanchard-Wrigglesworth, Edward and McGraw, Marie C and Cavallo, Steven M},
  journal={Journal of Climate},
  volume={35},
  number={1},
  pages={73--89},
  year={2022}
}

@article{cc4,
author = {Naud, Catherine M. and Martin, Jonathan E. and Ghosh, Poushali and Elsaesser, Gregory S. and Booth, James F. and Posselt, Derek J.},
title = {Lifecycle-Type Matters for Extratropical Cyclone Precipitation Production},
journal = {Geophysical Research Letters},
volume = {52},
number = {8},
pages = {e2025GL115153},
doi = {https://doi.org/10.1029/2025GL115153},
url = {https://agupubs.onlinelibrary.wiley.com/doi/abs/10.1029/2025GL115153},
eprint = {https://agupubs.onlinelibrary.wiley.com/doi/pdf/10.1029/2025GL115153},
note = {e2025GL115153 2025GL115153},
year = {2025}
}

@article{gertler2019changing,
  title={Changing available energy for extratropical cyclones and associated convection in Northern Hemisphere summer},
  author={Gertler, Charles G and O’Gorman, Paul A},
  journal={Proceedings of the National Academy of Sciences},
  volume={116},
  number={10},
  pages={4105--4110},
  year={2019},
  publisher={National Academy of Sciences}
}

@misc{shi2025deeplearningfoundationmodels,
      title={Deep Learning and Foundation Models for Weather Prediction: A Survey}, 
      author={Jimeng Shi and Azam Shirali and Bowen Jin and Sizhe Zhou and Wei Hu and Rahuul Rangaraj and Shaowen Wang and Jiawei Han and Zhaonan Wang and Upmanu Lall and Yanzhao Wu and Leonardo Bobadilla and Giri Narasimhan},
      year={2025},
      eprint={2501.06907},
      archivePrefix={arXiv},
      primaryClass={cs.LG},
      url={https://arxiv.org/abs/2501.06907}, 
}

@article{ho2020denoising,
  title={Denoising diffusion probabilistic models},
  author={Ho, Jonathan and Jain, Ajay and Abbeel, Pieter},
  journal={Advances in neural information processing systems},
  volume={33},
  pages={6840--6851},
  year={2020}
}

@article{ho2022video,
  title={Video diffusion models},
  author={Ho, Jonathan and Salimans, Tim and Gritsenko, Alexey and Chan, William and Norouzi, Mohammad and Fleet, David J},
  journal={Advances in neural information processing systems},
  volume={35},
  pages={8633--8646},
  year={2022}
}

@article{kullback1951information,
  title={On information and sufficiency},
  author={Kullback, Solomon and Leibler, Richard A},
  journal={The annals of mathematical statistics},
  volume={22},
  number={1},
  pages={79--86},
  year={1951},
  publisher={JSTOR}
}

@inproceedings{ronneberger2015u,
  title={U-net: Convolutional networks for biomedical image segmentation},
  author={Ronneberger, Olaf and Fischer, Philipp and Brox, Thomas},
  booktitle={International Conference on Medical image computing and computer-assisted intervention},
  pages={234--241},
  year={2015},
  organization={Springer}
}

@article{singer2022make,
  title={Make-a-video: Text-to-video generation without text-video data},
  author={Singer, Uriel and Polyak, Adam and Hayes, Thomas and Yin, Xi and An, Jie and Zhang, Songyang and Hu, Qiyuan and Yang, Harry and Ashual, Oron and Gafni, Oran and others},
  journal={arXiv preprint arXiv:2209.14792},
  year={2022}
}

@article{song2020score,
  title={Score-based generative modeling through stochastic differential equations},
  author={Song, Yang and Sohl-Dickstein, Jascha and Kingma, Diederik P and Kumar, Abhishek and Ermon, Stefano and Poole, Ben},
  journal={arXiv preprint arXiv:2011.13456},
  year={2020}
}

@article{hersbach2020era5,
  title={The ERA5 global reanalysis},
  author={Hersbach, Hans and Bell, Bill and Berrisford, Paul and Hirahara, Shoji and Hor{\'a}nyi, Andr{\'a}s and Mu{\~n}oz-Sabater, Joaqu{\'\i}n and Nicolas, Julien and Peubey, Carole and Radu, Raluca and Schepers, Dinand and others},
  journal={Quarterly journal of the royal meteorological society},
  volume={146},
  number={730},
  pages={1999--2049},
  year={2020},
  publisher={Wiley Online Library}
}

@article{nguyen2023climax,
  title={Climax: A foundation model for weather and climate},
  author={Nguyen, Tung and Brandstetter, Johannes and Kapoor, Ashish and Gupta, Jayesh K and Grover, Aditya},
  journal={arXiv preprint arXiv:2301.10343},
  year={2023}
}

@article{bodnar2024aurora,
  title={Aurora: A foundation model of the atmosphere},
  author={Bodnar, Cristian and Bruinsma, Wessel P and Lucic, Ana and Stanley, Megan and Brandstetter, Johannes and Garvan, Patrick and Riechert, Maik and Weyn, Jonathan and Dong, Haiyu and Vaughan, Anna and others},
  journal={arXiv preprint arXiv:2405.13063},
  volume={1},
  number={8},
  year={2024}
}

@inproceedings{shi2025codicast,
  title={CoDiCast: Conditional Diffusion Model for Global Weather Forecasting with Uncertainty Quantification},
  author={Shi, Jimeng and Jin, Bowen and Han, Jiawei and Gopalakrishnan, Sundararaman and Narasimhan, Giri},
  year={2025},
  organization={International Joint Conferences on Artificial Intelligence Organization}
}

@inproceedings{andrae2025continuous,
  title={Continuous ensemble weather forecasting with diffusion models},
  author={Andrae, Martin and Landelius, Tomas and Oskarsson, Joel and Lindsten, Fredrik},
  booktitle={International Conference on Learning Representations},
  volume={2025},
  pages={26392--26416},
  year={2025}
}

@article{rasp2020weatherbench,
  title={WeatherBench: a benchmark data set for data-driven weather forecasting},
  author={Rasp, Stephan and Dueben, Peter D and Scher, Sebastian and Weyn, Jonathan A and Mouatadid, Soukayna and Thuerey, Nils},
  journal={Journal of Advances in Modeling Earth Systems},
  volume={12},
  number={11},
  pages={e2020MS002203},
  year={2020},
  publisher={Wiley Online Library}
}

@inproceedings{hang2023efficient,
  title={Efficient diffusion training via min-snr weighting strategy},
  author={Hang, Tiankai and Gu, Shuyang and Li, Chen and Bao, Jianmin and Chen, Dong and Hu, Han and Geng, Xin and Guo, Baining},
  booktitle={Proceedings of the IEEE/CVF international conference on computer vision},
  pages={7441--7451},
  year={2023}
}

@article{https://doi.org/10.1002/2016RG000519,
author = {Catto, J. L.},
title = {Extratropical cyclone classification and its use in climate studies},
journal = {Reviews of Geophysics},
volume = {54},
number = {2},
pages = {486-520},
doi = {https://doi.org/10.1002/2016RG000519},
url = {https://agupubs.onlinelibrary.wiley.com/doi/abs/10.1002/2016RG000519},
eprint = {https://agupubs.onlinelibrary.wiley.com/doi/pdf/10.1002/2016RG000519},
year = {2016}
}

@mastersthesis{steiner2021regional,
  title={A Regional Comparison of Bomb Cyclones in the Central Plains and Western Atlantic},
  author={Steiner, Joshua C},
  year={2021},
  school={The Ohio State University}
}

@article{mardani2024residual,
  title={Residual diffusion modeling for km-scale atmospheric downscaling},
  author={Mardani, Morteza and Brenowitz, Noah and Cohen, Yair and Pathak, Jaideep and Chen, Chieh-Yu and Liu, Cheng-Chin and Vahdat, Arash and Kashinath, Karthik and Kautz, Jan and Pritchard, Mike},
  year={2024}
}

@article{andrychowicz2023deep,
  title={Deep learning for day forecasts from sparse observations},
  author={Andrychowicz, Marcin and Espeholt, Lasse and Li, Di and Merchant, Samier and Merose, Alexander and Zyda, Fred and Agrawal, Shreya and Kalchbrenner, Nal},
  journal={arXiv preprint arXiv:2306.06079},
  year={2023}
}

@article{schmude2024prithvi,
  title={Prithvi wxc: Foundation model for weather and climate},
  author={Schmude, Johannes and Roy, Sujit and Trojak, Will and Jakubik, Johannes and Civitarese, Daniel Salles and Singh, Shraddha and Kuehnert, Julian and Ankur, Kumar and Gupta, Aman and Phillips, Christopher E and others},
  journal={arXiv preprint arXiv:2409.13598},
  year={2024}
}

@article{gao2023prediff,
  title={Prediff: Precipitation nowcasting with latent diffusion models},
  author={Gao, Zhihan and Shi, Xingjian and Han, Boran and Wang, Hao and Jin, Xiaoyong and Maddix, Danielle and Zhu, Yi and Li, Mu and Wang, Yuyang Bernie},
  journal={Advances in Neural Information Processing Systems},
  volume={36},
  pages={78621--78656},
  year={2023}
}

@article{li2023seeds,
  title={Seeds: Emulation of weather forecast ensembles with diffusion models},
  author={Li, Lizao and Carver, Rob and Lopez-Gomez, Ignacio and Sha, Fei and Anderson, John},
  journal={arXiv preprint arXiv:2306.14066},
  year={2023}
}

@inproceedings{lugmayr2022repaint,
  title={Repaint: Inpainting using denoising diffusion probabilistic models},
  author={Lugmayr, Andreas and Danelljan, Martin and Romero, Andres and Yu, Fisher and Timofte, Radu and Van Gool, Luc},
  booktitle={Proceedings of the IEEE/CVF conference on computer vision and pattern recognition},
  pages={11461--11471},
  year={2022}
}

@article{lam2023learning,
  title={Learning skillful medium-range global weather forecasting},
  author={Lam, Remi and Sanchez-Gonzalez, Alvaro and Willson, Matthew and Wirnsberger, Peter and Fortunato, Meire and Alet, Ferran and Ravuri, Suman and Ewalds, Timo and Eaton-Rosen, Zach and Hu, Weihua and others},
  journal={Science},
  volume={382},
  number={6677},
  pages={1416--1421},
  year={2023},
  publisher={American Association for the Advancement of Science}
}

@article{price2023gencast,
  title={Gencast: Diffusion-based ensemble forecasting for medium-range weather},
  author={Price, Ilan and Sanchez-Gonzalez, Alvaro and Alet, Ferran and Andersson, Tom R and El-Kadi, Andrew and Masters, Dominic and Ewalds, Timo and Stott, Jacklynn and Mohamed, Shakir and Battaglia, Peter and others},
  journal={arXiv preprint arXiv:2312.15796},
  year={2023}
}

@article{kochkov2024neural,
  title={Neural general circulation models for weather and climate},
  author={Kochkov, Dmitrii and Yuval, Janni and Langmore, Ian and Norgaard, Peter and Smith, Jamie and Mooers, Griffin and Kl{\"o}wer, Milan and Lottes, James and Rasp, Stephan and D{\"u}ben, Peter and others},
  journal={Nature},
  volume={632},
  number={8027},
  pages={1060--1066},
  year={2024},
  publisher={Nature Publishing Group UK London}
}

@article{
doi:10.1126/science.aav9527,
author = {Ian R. Young  and Agustinus Ribal },
title = {Multiplatform evaluation of global trends in wind speed and wave height},
journal = {Science},
volume = {364},
number = {6440},
pages = {548-552},
year = {2019},
doi = {10.1126/science.aav9527},
URL = {https://www.science.org/doi/abs/10.1126/science.aav9527},
eprint = {https://www.science.org/doi/pdf/10.1126/science.aav9527}}

@book{martin2013mid,
  title={Mid-latitude atmospheric dynamics: a first course},
  author={Martin, Jonathan E},
  year={2013},
  publisher={John Wiley \& Sons}
}

@article{willoughby2006parametric,
  title={Parametric representation of the primary hurricane vortex. Part II: A new family of sectionally continuous profiles},
  author={Willoughby, Hugh E and Darling, RWR and Rahn, ME},
  journal={Monthly weather review},
  volume={134},
  number={4},
  pages={1102--1120},
  year={2006}
}

@article{zhang2026physics,
  title={Physics-based models outperform AI weather forecasts of record-breaking extremes},
  author={Zhang, Zhongwei and Fischer, Erich and Zscheischler, Jakob and Engelke, Sebastian},
  journal={Science Advances},
  volume={12},
  number={18},
  pages={eaec1433},
  year={2026},
  publisher={American Association for the Advancement of Science}
}

@article{snaiki2022knowledge,
  title={Knowledge-enhanced deep learning for simulation of extratropical cyclone wind risk},
  author={Snaiki, Reda and Wu, Teng},
  journal={Atmosphere},
  volume={13},
  number={5},
  pages={757},
  year={2022},
  publisher={MDPI}
}

@misc{blattmann_stable_2023,
	title = {Stable {Video} {Diffusion}: {Scaling} {Latent} {Video} {Diffusion} {Models} to {Large} {Datasets}},
	shorttitle = {Stable {Video} {Diffusion}},
	url = {http://arxiv.org/abs/2311.15127},
	doi = {10.48550/arXiv.2311.15127},
	urldate = {2025-03-18},
	publisher = {arXiv},
	author = {Blattmann, Andreas and Dockhorn, Tim and Kulal, Sumith and Mendelevitch, Daniel and Kilian, Maciej and Lorenz, Dominik and Levi, Yam and English, Zion and Voleti, Vikram and Letts, Adam and Jampani, Varun and Rombach, Robin},
	month = nov,
	year = {2023},
	note = {arXiv:2311.15127 [cs]},
}

@article{booth_extratropical_2018,
	title = {Extratropical {Cyclone} {Precipitation} {Life} {Cycles}: {A} {Satellite}-{Based} {Analysis}},
	volume = {45},
	copyright = {©2018. American Geophysical Union. All Rights Reserved.},
	issn = {1944-8007},
	shorttitle = {Extratropical {Cyclone} {Precipitation} {Life} {Cycles}},
	url = {https://onlinelibrary.wiley.com/doi/abs/10.1029/2018GL078977},
	doi = {10.1029/2018GL078977},
	language = {en},
	number = {16},
	urldate = {2025-02-15},
	journal = {Geophysical Research Letters},
	author = {Booth, James F. and Naud, Catherine M. and Jeyaratnam, Jeyavinoth},
	year = {2018},
	note = {\_eprint: https://onlinelibrary.wiley.com/doi/pdf/10.1029/2018GL078977},
	pages = {8647--8654},
}

@article{hawcroft_how_2012,
	title = {How much {Northern} {Hemisphere} precipitation is associated with extratropical cyclones?},
	volume = {39},
	issn = {1944-8007},
	url = {https://onlinelibrary.wiley.com/doi/abs/10.1029/2012GL053866},
	doi = {10.1029/2012GL053866},
	language = {en},
	number = {24},
	urldate = {2024-11-20},
	journal = {Geophysical Research Letters},
	author = {Hawcroft, M. K. and Shaffrey, L. C. and Hodges, K. I. and Dacre, H. F.},
	year = {2012},
	note = {\_eprint: https://onlinelibrary.wiley.com/doi/pdf/10.1029/2012GL053866},
}

@article{attinger_systematic_2021,
	title = {Systematic assessment of the diabatic processes that modify low-level potential vorticity in extratropical cyclones},
	volume = {2},
	doi = {10.5194/wcd-2-1073-2021},
	journal = {Weather and Climate Dynamics},
	author = {Attinger, Roman and Spreitzer, Elisa and Boettcher, Maxi and Wernli, Heini and Joos, Hanna},
	month = nov,
	year = {2021},
	pages = {1073--1091},
}
\bibliographystyle{tmlr}

\appendix

\section{Glossary of Terms and Abbreviations} \label{app:glossary}

We include definitions for common abbreviations in this section, as well as important technical terms.

\begin{itemize}[topsep=0pt,itemsep=-0.08cm,leftmargin=*,partopsep=0pt, parsep=4pt]
    \item \textbf{ACC (Anomaly Correlation Coefficient):} A standard meteorological metric that measures the spatial pattern correlation of predicted fields to evaluate spatial and navigational accuracy \citep{wilks2011statistical}.
    \item \textbf{Autoregressive Synthesis:} A generation approach where the model predicts each subsequent timestep based on the previous one(s).
    \item \textbf{CEF (Continuous Ensemble Forecasting):} A diffusion-based global forecasting ensemble that operates via a continuous deterministic ODE solver \citep{andrae2025continuous}.
    \item \textbf{DDPM (Denoising Diffusion Probabilistic Model):} A generative machine learning framework that synthesizes data by learning to reverse a Markovian forward diffusion process that progressively adds Gaussian noise to a sample, see Appendix~\ref{app:dm-background} \citep{ho2020denoising}.
    \item \textbf{ERA5}: The fifth generation ECMWF atmospheric reanalysis of the global climate, used as the primary data source for training and evaluating the models in this study \citep{hersbach2020era5}.
    \item \textbf{Extratropical Cyclone}: An evolving, mid-latitude weather system driven by atmospheric temperature gradients, associated with strong wind, heavy precipitation and large-scale energy transfer \citep{martin2013mid}.
    \item \textbf{Fine-tuning:} A machine learning procedure where a pre-trained model is further trained on a task-specific dataset (such as climate fields) to adapt its generalized capabilities to a specialized domain.
    \item \textbf{FBI (Frequency Bias Index):} A metric evaluating extreme intensity, calculated as the ratio of predicted extreme wind pixels ($>20$ m/s) to actual extreme wind pixels \citep{wilks2011statistical}.
    \item \textbf{FSS (Fractions Skill Score):} A neighborhood-based spatial verification metric that measures the spatial overlap accuracy specifically for extreme ($>20$ m/s) wind distributions \citep{roberts2008scale}.
    \item \textbf{GeoDES (Geospatial Diffusion-based Evolution Synthesis):} The primary architecture proposed in this work; a storm-centered, spatiotemporal diffusion model designed for synthesizing large-scale storm structure evolution.
    \item \textbf{Hallucination:} An undesirable effect of generative model outputs appearing structurally coherent, but physically incorrect or purely fabricated, often failing to respect underlying scientific laws.
    \item \textbf{Latent Space:} A compressed, low-dimensional mathematical representation of data where a model encodes key features, allowing for efficient processing and synthesis of complex spatial structures.
    \item \textbf{MCMS (Modeling, Analysis and Prediction Climatology of Mid-latitude Storminess):} Codebase and methodology used to identify and track the coordinates of extratropical cyclones within the global ERA5 dataset, see Appendix~\ref{app:preprocess-data}.
    \item \textbf{MSE (Mean Squared Error):} Standard loss function used during training, measuring the pixel-wise difference between ground-truth and predicted atmospheric states \citep{ho2022video}.
    \item \textbf{Non-Autoregressive Synthesis}: A generation approach where the model synthesizes an entire sequence of data (like a time-series of storm frames) simultaneously rather than step-by-step. This contrasts with autoregressive synthesis.
    \item \textbf{RMSE (Root Mean Square Error):} A standard pixel-wise evaluation metric calculated across all un-normalized atmospheric variables \citep{roberts2008scale}.
    \item \textbf{SACC (Synoptic Anomaly Correlation Coefficient):} An adaptation of the ACC metric that applies a spatial Gaussian filter prior to correlation, mitigating the double-penalty effect for structurally sound but slightly displaced storm predictions \citep{wilks2011statistical}.
    \item \textbf{SLP (Sea-Level Pressure):} One of the five primary atmospheric variables analyzed in this study; the minimum SLP is tracked to determine the center coordinates of the cyclonic storms.
    \item \textbf{SNR (Signal-to-Noise Ratio):} A weighting mechanism applied to the reverse denoising training objective to prioritize physical accuracy and stabilize intermediate noise scales \citep{hang2023efficient}.
    \item \textbf{SVD (Stable Video Diffusion):} An off-the-shelf image-to-video diffusion baseline model evaluated for its zero-shot physical viability in storm synthesis \citep{blattmann_stable_2023}.
    \item \textbf{Transformer:} A prominent neural network architecture that relies on attention mechanisms to capture long-range dependencies in data; widely used as the backbone for modern foundation models.
    \item \textbf{U-Net:} A convolutional neural network architecture commonly used in diffusion models, featuring a symmetric encoder-decoder structure that preserves spatial resolution for tasks like storm synthesis.
    \item \textbf{WFM (Weather Foundation Model):} Large-scale machine learning architectures (such as ClimaX and Aurora) pretrained on vast amounts of atmospheric data to perform generalized weather tasks.
    \item \textbf{Zero-shot Evaluation:} An evaluation paradigm where a machine learning model is tested on a task or dataset it was not explicitly trained for, demonstrating its ability to generalize to unseen scenarios without parameter updates.
\end{itemize}

\section{Background on Diffusion Modeling}\label{app:dm-background}
A Diffusion Model (DM) is capable of generating synthetic datapoints by learning a mapping from pure Gaussian noise to the desired data distribution \citep{ho2020denoising}. Denote a training dataset by $D\subseteq \mathbb{R}^d$. The \textit{forward diffusion process} is responsible for transforming a real datapoint $\mathbf x_0\in D$ with distribution $q(\mathbf x_0)$ into a tensor of Gaussian noise $\mathbf x_N$ over $N$ diffusion steps. This process is represented by a Markov chain and acts according to a variance schedule $\{\beta_1, \beta_2, \ldots, \beta_n\}$, to produce
\begin{align}\label{eq:dm-forward-open}
    q(\mathbf x_n | \mathbf x_{n-1}) &= \mathcal N(\mathbf x_n; \sqrt{1-\beta_n}\mathbf x_{n-1}, \beta_n \mathbf I) \text{ and}\\
    q(\mathbf x_{1:N}|\mathbf x_0) &= \prod_{n=1}^N q(\mathbf x_n | \mathbf x_{n-1}), n\in [1,N].
\end{align}
The forward process can also be described in the closed form across steps as $$q(\mathbf x_n | \mathbf x_0)=\mathcal N (\mathbf x_n; \sqrt{\bar{\alpha}_n}\mathbf x_0, (1-\bar{\alpha}_n)\mathbf I),$$
where $\alpha_n = 1-\beta_n$ and $\bar{\alpha}_n=\prod_{s=1}^n \alpha_s$. 

In the \textit{reverse denoising process}, the denoiser network uses $\mathbf x_N$ to predict a denoised version of $\mathbf x_0$ in steps. Given that $p_\Theta(\mathbf x_n)$ is the distribution at noise step $n$ parameterized by the denoiser $\Theta$, this process is performed as
\begin{equation}\label{eq:dm-reverse}
    p_\Theta(\mathbf x_{0:N}) = p(\mathbf x_N) \prod_{n=1}^N p_\Theta(\mathbf x_{n-1}|\mathbf x_n).
\end{equation}

For each noise step $n \in [1,N]$, the training process aims to minimize the Kullback-Leibler (KL) Divergence \citep{kullback1951information} between the predicted and ground-truth sample distribution at noise step $n-1$,
$$\mathcal L_n = D_{KL}(q(\mathbf x_{n-1}|\mathbf x_n) || p_\Theta(\mathbf x_{n-1}|\mathbf x_n)).$$
The term $q(\mathbf x_{n-1}|\mathbf x_n)$ is computed from $q(\mathbf x_{n}|\mathbf x_{n-1})$ using Bayes theorem and the open-form equation for the forward process. Note that in practice, most diffusion implementations (including GeoDES) use mean squared error (MSE) loss in place of KL Divergence, for computational efficiency and to encourage sharper, more detailed outputs \citep{ho2022video}. 

\section{Experimental Setup}

In this appendix, we provide further information on the setups for the experiments performed in Section~\ref{sec:evals}. Appendix~\ref{app:preprocess-data} includes information on how non-synthetic data, as well as model outputs, were processed, while Appendix~\ref{app:hyperparams} describes model hyperparameters. Appendix~\ref{app:compute} provides details on how computational requirements were determined as presented in Section~\ref{sec:exp3}.

\subsection{Data Processing Details}\label{app:preprocess-data}
\textbf{Storm selection:} Extratropical cyclones are identified within the ERA5 Sea Level Pressure 1940-2024 6-hourly global data on a $0.25^\circ\times 0.25^\circ$ grid using the NASA Modeling, Analysis and Prediction Climatology of Mid-latitude Storminess (MCMS) codebase\footnote{https://github.com/jfbooth/MIKE\_BAUERS\_MCMSV4}. For each extratropical cyclone, the codebase returns a list of timeframes in which the cyclone occurs, as well as the latitude and longitude of the storm's low-pressure center at each timeframe. We filter to only include storms that begin equatorward of $70^\circ$ and that persist for at least $8$ timeframes. For storms longer than $8$ timepoints, we only include the first $8$ timeframes. Next, we use a region mask to categorize and retain storms that occur in the North/South Atlantic and Pacific basins.

\textbf{Global models}\\
\textit{Train-time procedure:} Global model codebases are designed to expect continuous 6-hourly global data over a given period. Accordingly, models are trained on 6-hourly data over our entire training period (January 1940-August 2015).

\textit{Test-time procedure:} We sample each model using a 7-step lead time. For each storm in our test set (January 2016-December 2024), we select the model output that is prompted from the timestep that the storm begins. Beginning at the ground-truth initial storm location, we then track the storm through the synthetic global forecast in subsequent timesteps by locating the minimum Sea Level Pressure (SLP) within a $10^{\circ} \times 10^{\circ}$ bounding box centered on the previous frame's location. At each timestep, we then extract a $1600\times1600$km equal-area bounding box, centered at the storm's minimum SLP location. We perform interpolation as necessary to reach a $32\times32\times V$ tensor, where $V$ is the number of climate variables.

\textit{ClimaX SLP Derivation:} Because pretrained ClimaX does not offer SLP, we instead condition on and predict geopotential at 925 hPa and 2-meter temperature. The ClimaX-predicted SLP is then derived via the hypsometric equation,
$$P_{slp} = \ell e^{\Phi / (R_d T_{2m})},$$ 
where $\ell$ is the reference pressure level (925 hPa), $\Phi$ is the predicted geopotential, $R_d$ is the specific gas constant for dry air and $T_{2m}$ is 2-meter temperature. Consequently, storm tracking for ClimaX relies on geopotential minima in place of SLP.

\textbf{Patch-based models (GeoDES and SVD)}\\
\textit{Train-time procedure:} Using the ground-truth storm location data created by MCMS for train-time storms in a given basin of interest, we extract bounding boxes around each storm at each timestep of its lifecycle, centered on its minimum SLP location. Interpolating as needed, each storm is represented as a $32\times32\times V$ tensor, where the first two dimensions represent grid area and $V$ is the number of climate variables ($V=5$ in our experiments). GeoDES and SVD are trained on these samples directly.

\textit{Test-time procedure:} Forming ground-truth test set datapoints in an analogous fashion to the GeoDES and SVD train set datapoints above, both models are then prompted with the $0$-th timeframe of each storm in the test set. 

\textbf{Evaluations}\\
Given the storm-centered bounding boxes created by or extracted from each method, evaluations are performed against the ground-truth test set storm-centered bounding boxes created during the GeoDES/SVD test-time procedure.

\subsection{Hyperparameters}~\label{app:hyperparams}

\textbf{GeoDES:} GeoDES was implemented in Python 3.11.11 with Diffusers 0.35.0 and PyTorch 2.6.0. U-Net block out channels are set to $[512, 1024, 2048]$ and we leave the layers per block and cross-attention dimension to the diffusers library defaults of $2$ and $768$, respectively. 

We train GeoDES for a maximum total of 40 epochs (20 at 2D phase and 20 at 3D phase). The 2D and 3D phases are both performed with a patience of $5$ epochs, with patience determined according to the training loss. The learning rate is $1\mathrm{e}{-5}$ at the start of both 2D and 3D phases, selected from a search over $\{1\mathrm{e}{-3}, 1\mathrm{e}{-4}, 1\mathrm{e}{-5}, 1\mathrm{e}{-6}, 1\mathrm{e}{-7}, 1\mathrm{e}{-8}\}$. We perform the learning rate search at the 2D phase, select the two learning rates with lowest losses ($1\mathrm{e}{-4}$ and $1\mathrm{e}{-5}$), then perform the 3D loss search using both of these 2D settings to select the 3D model with lowest overall loss. Both 2D and 3D phases use the diffusers cosine schedule learning rate decay implementations with no warmup. We use MSE loss function and SNR of $5$.  

The training phase uses an AdamW optimizer and a DDPM scheduler with $1000$ timesteps. Our batch size is $1$, with $8$ gradient accumulation steps and no weight decay. Model weights are updated directly without the use of an Exponential Moving Average (EMA).

At the 3D phase, we train and sample with BFloat16 (bf16) mixed precision and use temporal noise correlation $\rho=0.95$ (see Section~\ref{sec:method-model} for details). At sample time, we use the DDIM sampler with $50$ timesteps and $\eta=1.0$ (we find that $\eta=0.0$ also produces viable, but more conservative, predictions).

\textbf{Baselines:} For the prior work baselines (CoDiCast, CEF, ClimaX, ClimaX FT, Aurora), we use the highest-resolution model available, with the hyperparameters selected in their respective codebases. We use BFloat 16 mixed precision fine-tuning for ClimaX FT. For SVD, we use the same hyperparameter search protocol as the GeoDES model to reach a learning rate of $1\mathrm{e}{-4}$.

\subsection{Details on Computational Requirements Estimation}~\label{app:compute}

\textbf{Estimation of Aurora FLOPs:} Because Aurora operates on global $0.25^\circ$ inputs, empirical FLOP calculation requires prohibitive amounts of system RAM and we instead estimate the FLOPs for this model. 

Aurora is a 1.25B-parameter transformer-based model, with inputs of dimensions $721\times1440$ per variable. A Perceiver encoder model performs 3D convolutions to extract $4\times4$ patches, compressing the grid to $180\times360=64800$ tokens per level.  Vertically, the encoder cross-attends the inputs to a latent depth of $Z=4$ levels (one surface and three atmospheric), regardless of the number of input variables or atmospheric levels\footnote{According to the official Microsoft Aurora implementation, https://github.com/microsoft/aurora}. Accordingly, the encoded input contains $4*64800=259200$ tokens. 

According to \cite{kaplan2020scalinglawsneurallanguage}, the cost of a single transformer inference-time forward pass may be estimated as $2PN$ for a $P$-parameter model and $N$ input tokens. Accordingly, each Aurora forward pass is approximately $2*(1.25\times10^9)*(2.59\times10^6)\approx 651$ TFLOPs. Multiplying by $7$ inference timesteps to achieve the $48$-hour forecasts within Section~\ref{sec:exp1} yields a total approximation of $4557$ TFLOPs.
% 1256000*259200*2= 651,110,400,000

We validate this FLOP estimate by calculating the corresponding Model FLOPs Utilization (MFU, \cite{chowdhery2022palmscalinglanguagemodeling}). As stated in Table~\ref{tab:compute}, we observe a $47.6$-second wall-clock latency for Aurora, which indicates a rate of $95.74$ TFLOP/second. These evaluations were performed on an NVIDIA RTX A6000, with a dense FP16/BF16 Tensor Core peak throughput of $154.8$ TFLOP/second. As such, the MFU is calculated as $\frac{\text{Model TFLOP/second}}{\text{Theoretical Peak Hardware TFLOP/second}} \approx 61.8\%$. Because Aurora performs a dense volumetric forward pass over a $259200$ token sequence at each autoregressive step, the computation maintains high arithmetic intensity. As such, the observed MFU aligns well with the established real-world vs theoretical upper bounds of dense PyTorch Transformer utilization \citep{hagemann2024efficientparallelizationlayoutslargescale, pope2023efficiently}.

\textbf{Estimation of Saved Compute Using 2D vs 3D GeoDES Training:} The 3D model uses $123,662$ GFLOPs to produce all $8$ frames of a single datapoint. The 2D model uses $1,182$ GFLOPs to produce a single (non-temporal) frame, or $9,455$ GFLOPs to produce $8$ frames. As such, we estimate the saved compute:
$$\frac{1}{2}+ \frac{1}{2}\cdot \frac{9455}{123662}=0.538.$$

\section{Additional Storm Modeling Realism Results}\label{app:moreresults}

We now present additional analyses of our main paper experiments presented in Section~\ref{sec:evals}. Appendix~\ref{app:exp1-plots} provides visualizations to accompany our main results on generative storm realism in Section~\ref{sec:exp1}, while Appendix~\ref{app:altmetrics} contains additional performance metrics (average and per-variable RMSE, synoptic and non-synoptic ACC) for the experiments presented in Sections~\ref{sec:exp1}, \ref{sec:exp2} and \ref{sec:exp4}

\subsection{Supplemental Plots and Visualizations}\label{app:exp1-plots}

In this section, we provide additional figures to visualize the performance of each method from Section~\ref{sec:exp1}. For explanations of each method, please refer to Section~\ref{sec:evals-setup}.

\begin{figure}[h]
    \centering
    \includegraphics[width=0.95\textwidth]{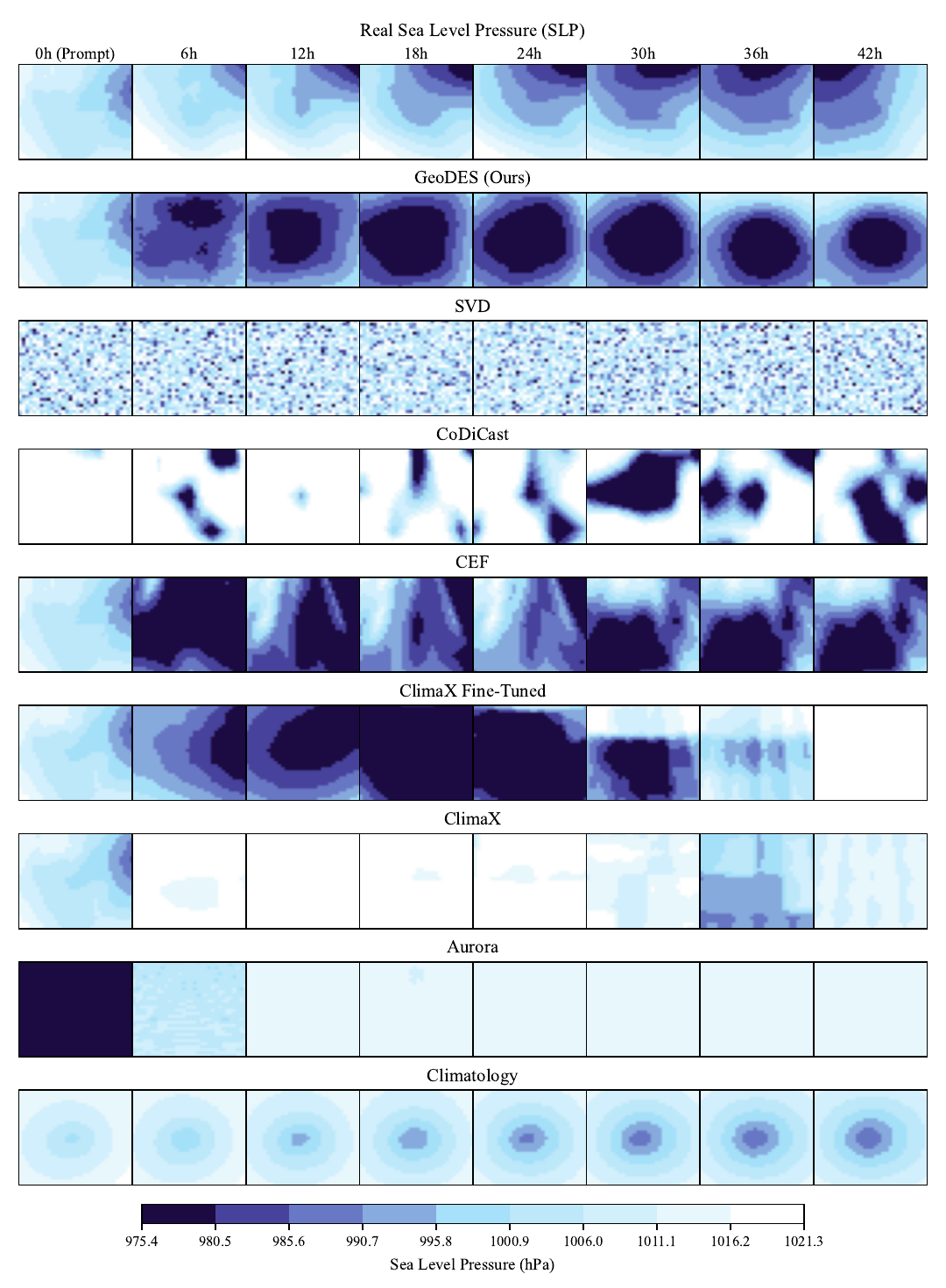}
    \vspace{-0.5cm}
    \caption{Sea Level Pressure (SLP) of the extratropical cyclone in Figure~\ref{fig:compare} and corresponding predictions from each synthesis method. While global WFMs correctly identify the macroscopic background pressure gradient, they fail to resolve the deep, localized minimum of the storm core. Conversely, GeoDES perfectly captures the tight pressure gradients and depth of the localized cyclone core.}
    \label{fig:compare_slp}
\end{figure}

\begin{figure}[h]
    \centering
    \includegraphics[width=0.95\textwidth]{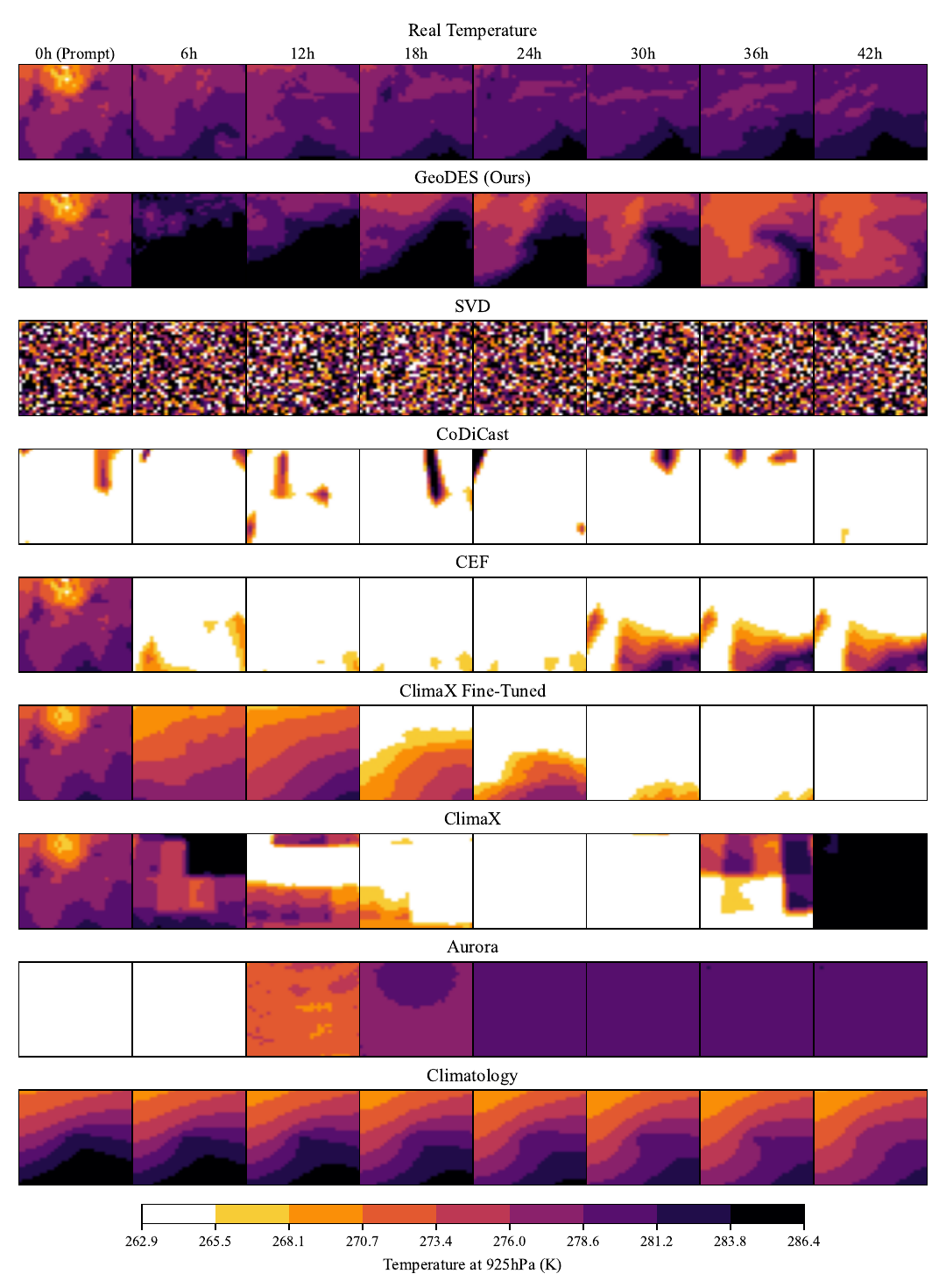}
    \vspace{-0.5cm}
    \caption{Temperature at 925hPa of the extratropical cyclone in Figure~\ref{fig:compare}. Extratropical cyclones are fundamentally driven by the interaction of contrasting air masses. While baseline models act as spatial low-pass filters that blur these boundaries, GeoDES successfully preserves the sharp thermodynamic gradients of the moving weather fronts and the swirling thermal advection of the storm over the 42-hour rollout.}
    \label{fig:compare_temperature}
\end{figure}

\begin{figure}[h]
    \centering
    \includegraphics[width=0.95\textwidth]{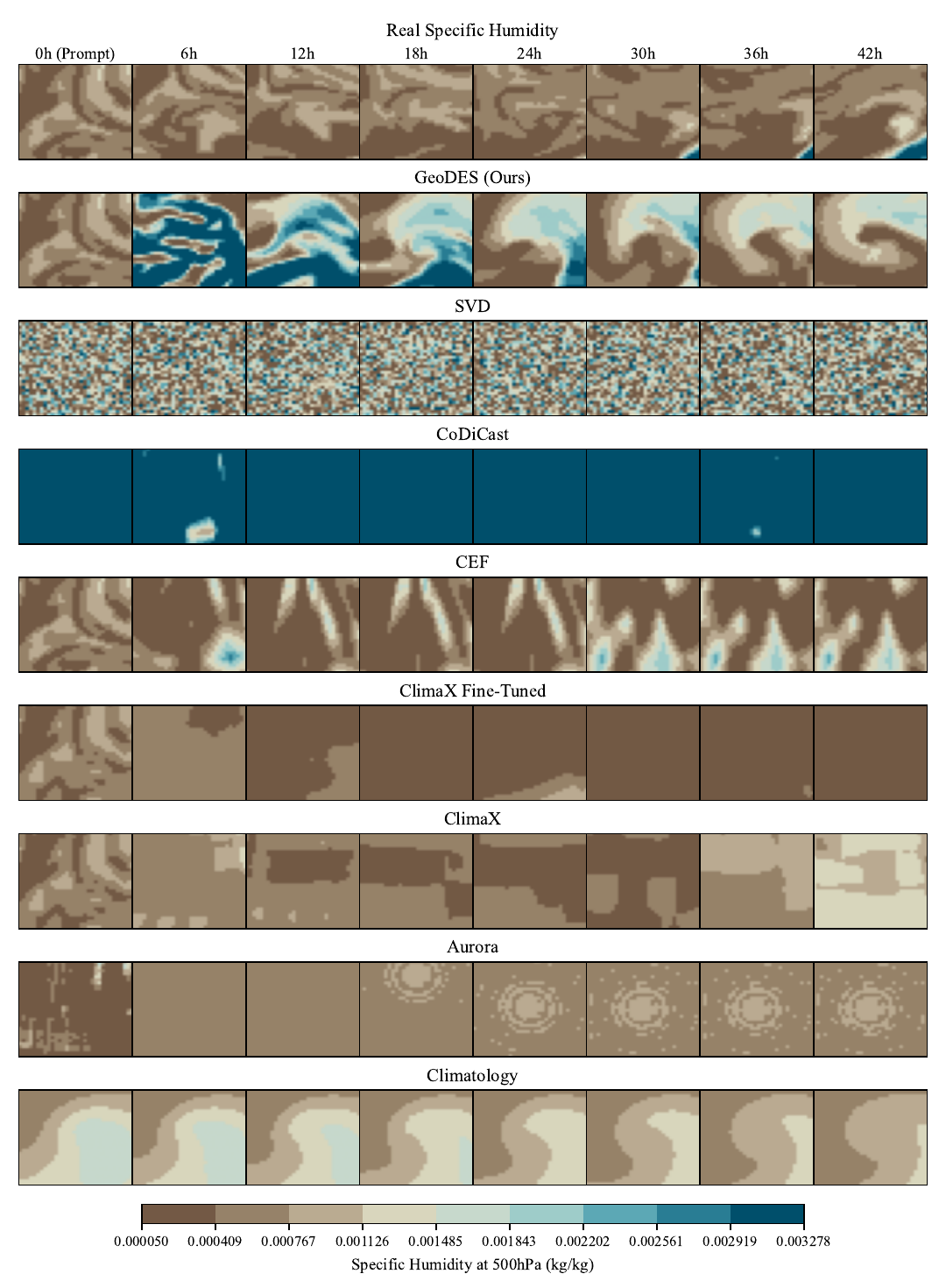}
    \vspace{-0.5cm}
    \caption{Specific Humidity at 500hPa of the extratropical cyclone in Figure~\ref{fig:compare}. Similar to the temperature fields, GeoDES accurately maintains the structural integrity and sharp gradients of the moisture fronts, avoiding the catastrophic spatial smoothing exhibited by the global baselines and the unconstrained noise injection of CoDiCast and SVD.}
    \label{fig:compare_humidity}
\end{figure}

\textbf{Example storm outputs:} We visualize the sea-level pressure, temperature and humidity outputs of each method compared to a real storm in Figures~\ref{fig:compare_slp}, \ref{fig:compare_temperature}, \ref{fig:compare_humidity}, respectively. The wind magnitude outputs for the same storm across all methods are presented in Figure~\ref{fig:compare}. 

Together, these figures support the narrative told by the performance metrics in Table~\ref{tab:exp-1}. Global Weather Foundation Models (ClimaX, Aurora) and other global autoregressive models (CEF) suffer from catastrophic smoothing. They fail to resolve the sharp thermodynamic gradients of moving weather fronts (visible in humidity and temperature) and wash out the high-frequency energy bands of the wind field. Conversely, the off-the-shelf video diffusion baseline (SVD) generates high-frequency energy but lacks physical constraints, collapsing into unorganized visual noise. CoDiCast, while not as severe as SVD, presents a similar high-energy failure mode, with overly energetic wind field and steep spatial gradients in its SLP, temperature and humidity predictions. GeoDES is the only architecture capable of preserving both the sharp frontal boundaries and the organized, high-frequency turbulence of the cyclone, balancing structural realism with intensity.

% compare SLP

% compare temperature

% compare humidity

\clearpage

% psd 
\begin{figure}[htbp]
    \centering
    
    % --- First Row ---
    \begin{subfigure}[b]{0.24\textwidth}
        \centering
        \includegraphics[width=\textwidth]{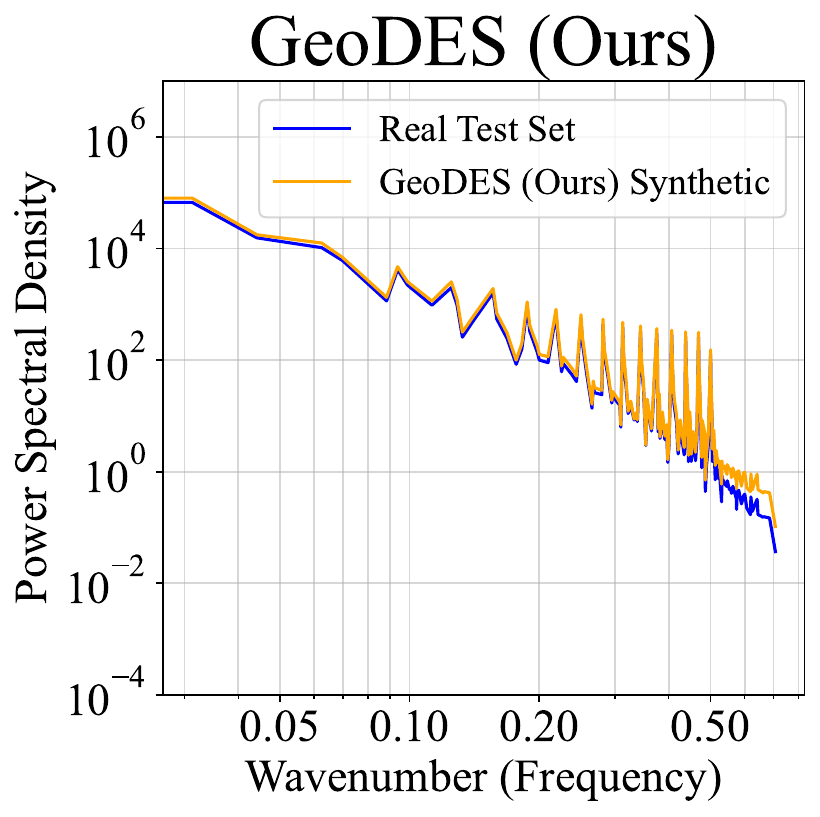}
        \label{fig:sub1}
    \end{subfigure}
    \hfill
    \begin{subfigure}[b]{0.24\textwidth}
        \centering
        \includegraphics[width=\textwidth]{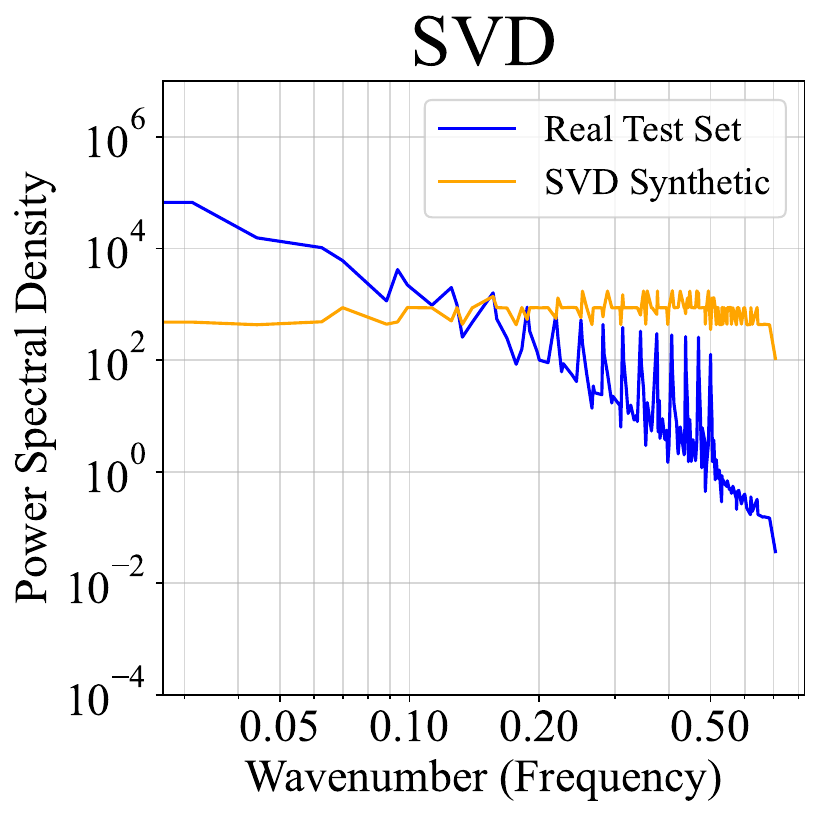}
        \label{fig:sub2}
    \end{subfigure}
    \hfill
    \begin{subfigure}[b]{0.24\textwidth}
        \centering
        \includegraphics[width=\textwidth]{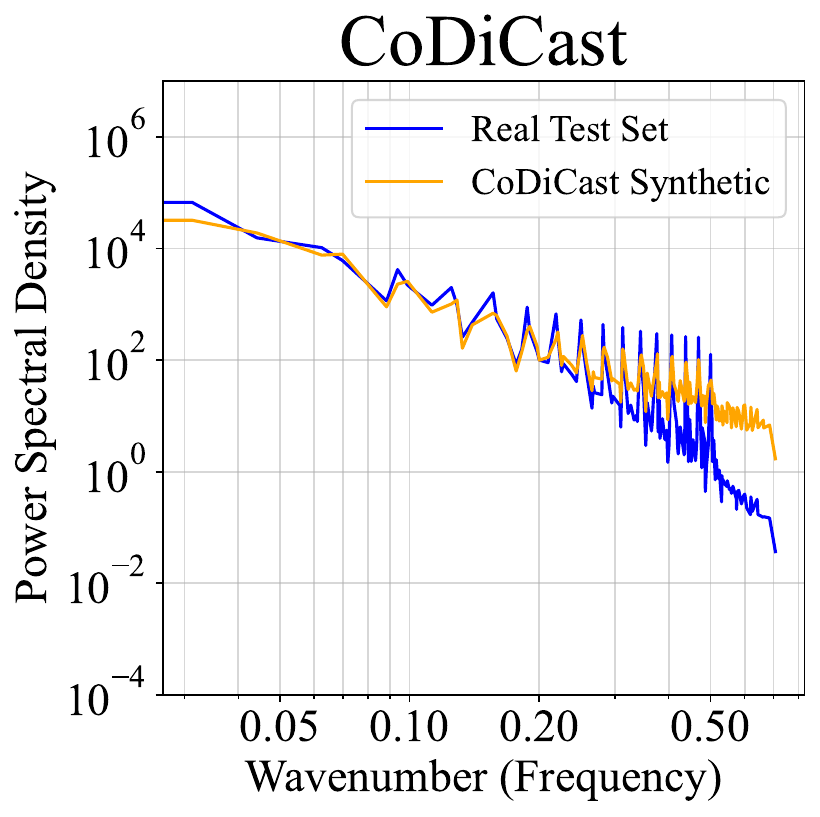}
        \label{fig:sub3}
    \end{subfigure}
    \hfill
    \begin{subfigure}[b]{0.24\textwidth}
        \centering
        \includegraphics[width=\textwidth]{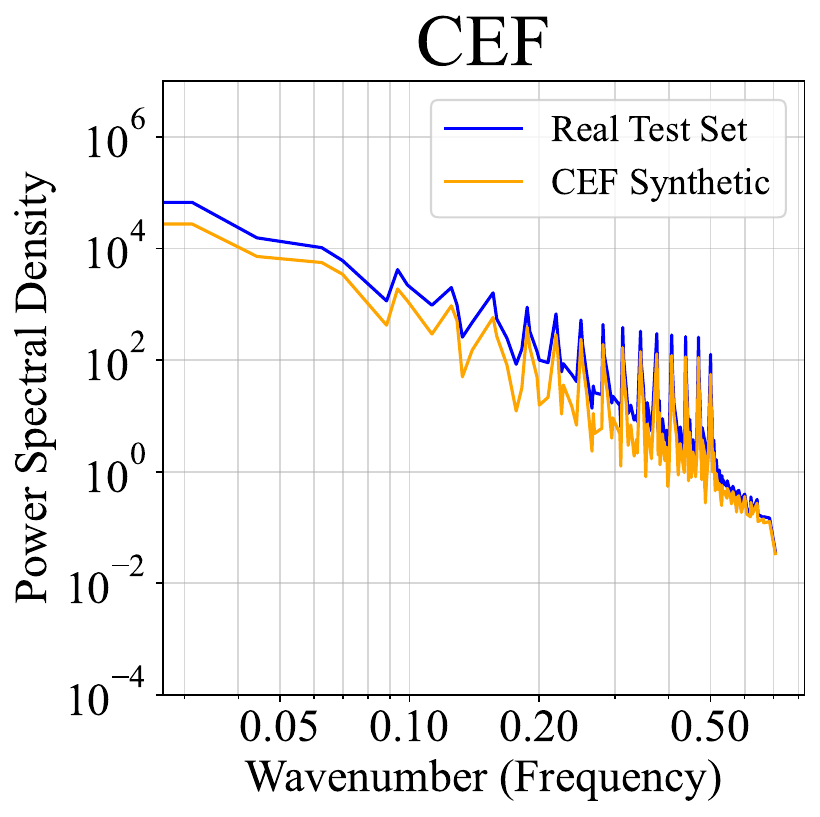}
        \label{fig:sub4}
    \end{subfigure}

    % --- Second Row ---
    \begin{subfigure}[b]{0.24\textwidth}
        \centering
        \includegraphics[width=\textwidth]{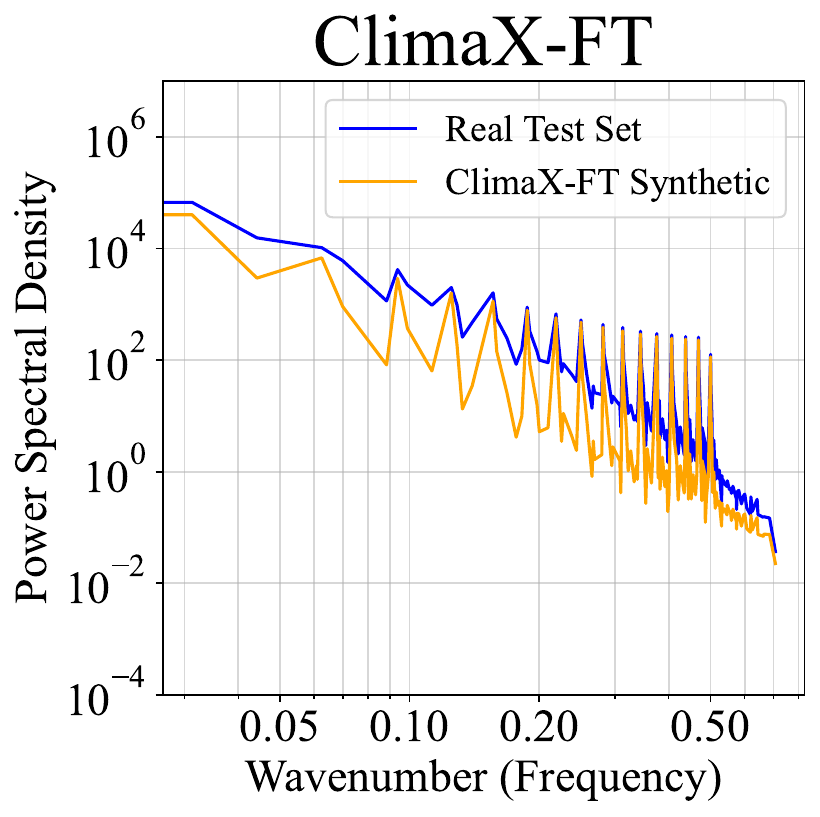}
        \label{fig:sub5}
    \end{subfigure}
    \hfill
    \begin{subfigure}[b]{0.24\textwidth}
        \centering
        \includegraphics[width=\textwidth]{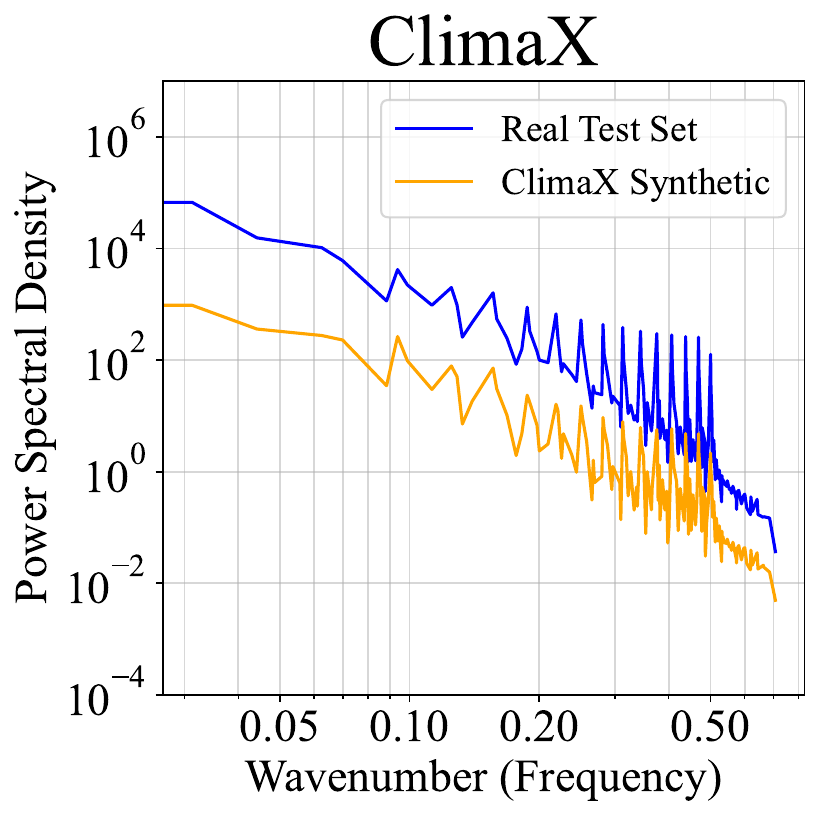}
        \label{fig:sub6}
    \end{subfigure}
    \hfill
    \begin{subfigure}[b]{0.24\textwidth}
        \centering
        \includegraphics[width=\textwidth]{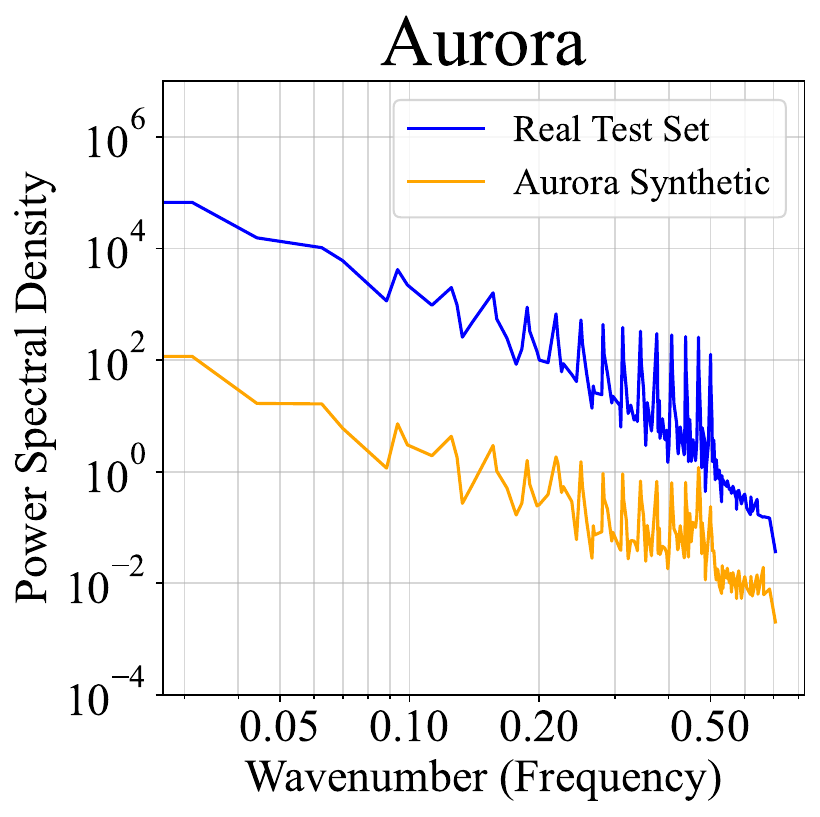}
        \label{fig:sub7}
    \end{subfigure}
    \hfill
    \begin{subfigure}[b]{0.24\textwidth}
        \centering
        \includegraphics[width=\textwidth]{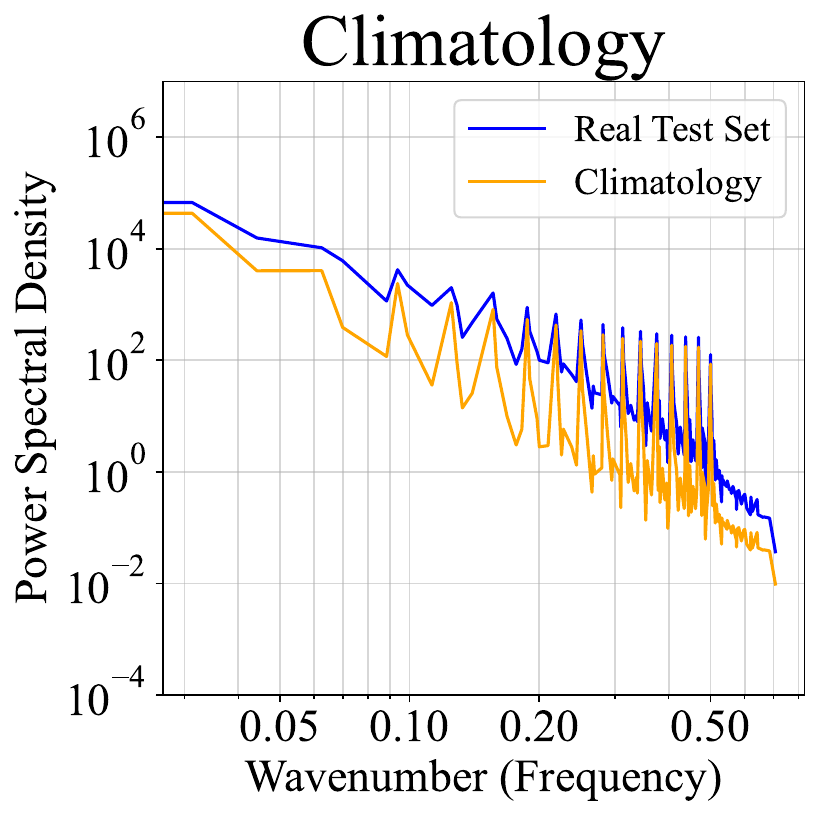}
        \label{fig:sub8}
    \end{subfigure}
% \vspace{-0.8cm}
    \caption{Power Spectral Density (PSD) of Wind Magnitude, visualizing how kinetic energy is distributed across spatial scales from macro-scale (left) to fine-grained, localized structures (right). Global models (ClimaX, Aurora) fall below the ground-truth test set (blue), indicating spatial smoothing. SVD and CoDiCast diverge above the ground truth at high frequencies, indicating noise hallucination. GeoDES uniquely preserves the kinetic energy distribution across spatial scales.}
    \label{fig:psd}
\end{figure}

\textbf{Power Spectral Density (PSD) Graphs:} Figure~\ref{fig:psd} contains one PSD graph per method, with the synthesis method represented by the orange line and the non-synthetic test set represented by the blue line. A PSD plot visualizes how kinetic energy is distributed across different spatial scales within a storm. The x-axis represents spatial frequency (wavenumber), transitioning from large, macro-scale weather patterns on the left to fine-grained, highly localized storm structures on the right. A physically realistic model will produce a synthetic energy spectrum that tightly tracks the ground-truth observational spectrum.

Most models (CEF, ClimaX FT, ClimaX and Aurora) have PSD lines that fall consistently beneath the test set PSD line, resembling the climatology baseline PSD. This behavior is indicative of overly smooth predictions and a failure to generate fine grained storm structures, which agrees with the general failure mode of these models as described in Section~\ref{sec:exp1} according to Table~\ref{tab:exp-1}. Meanwhile, SVD's and CoDiCast's PSD lines both fall above the test set line in the right half of their respective plots, which represents an over-prediction of high-energy features. While GeoDES presents a slight over-estimation bias in the highest-frequency features, across the full range of spatial frequencies, \textit{GeoDES is the model that tracks the closest to the non-synthetic test set.}

\begin{figure}[h]
    \centering
    \includegraphics[width=\textwidth]{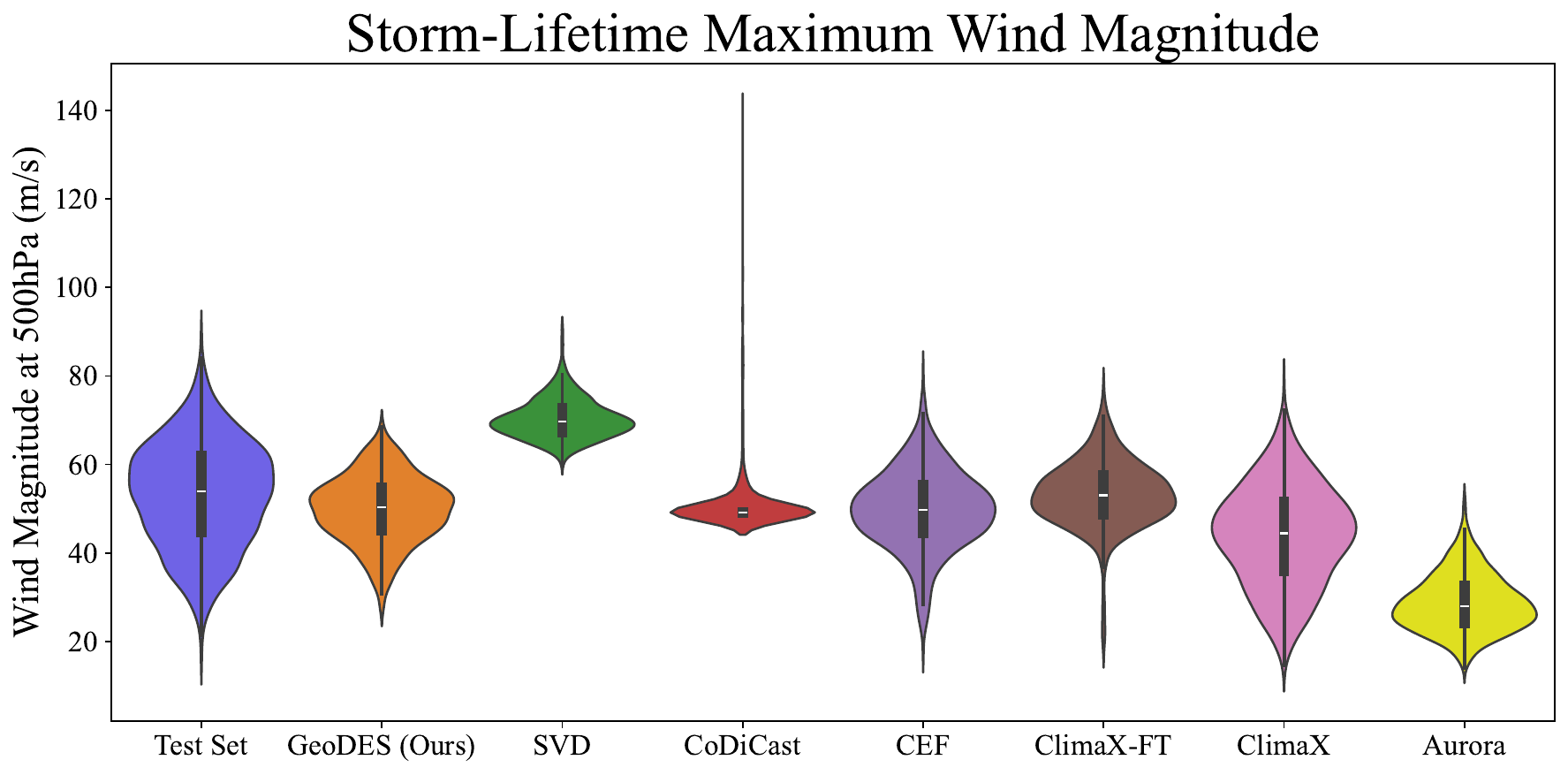}
    \caption{Violin plots detailing the distributions of storm-lifetime maximum wind speeds across the North Atlantic test set. While GeoDES successfully recovers the core distribution without succumbing to the unphysical runaway hallucinations of autoregressive baselines (e.g., CoDiCast), its maximum tail is bounded at roughly $75$ m/s due to the 99th-percentile clipping enforced by the hybrid normalizer. This clipping prevents the undesirable outlier effects demonstrated by SVD and CoDiCast.}
    \label{fig:violin_windmag}
\end{figure}

\begin{figure}[h]
    \centering
    \includegraphics[width=\textwidth]{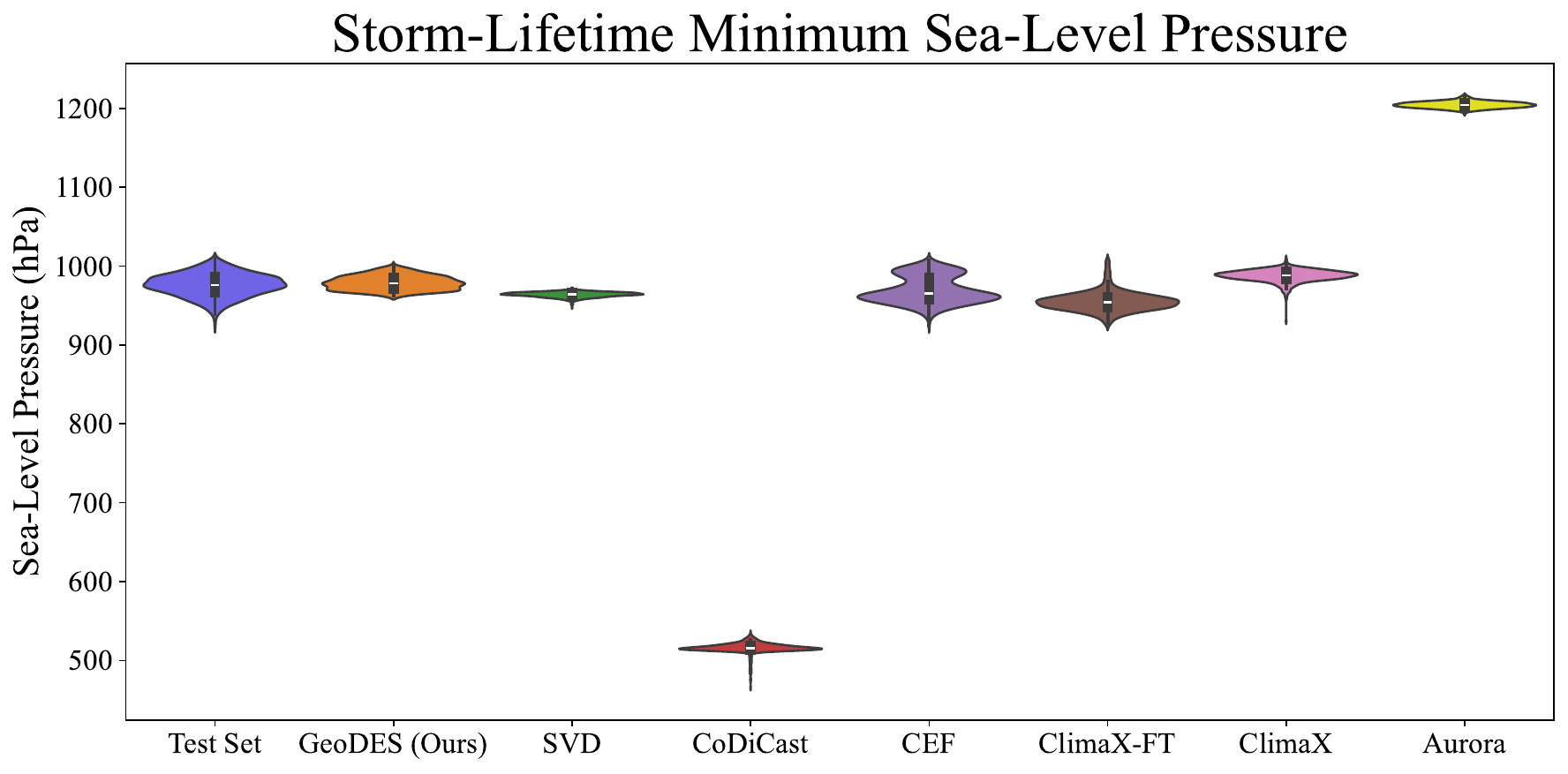}
    \caption{Distributions of storm-lifetime minimum Sea-Level Pressure (SLP) across the North Atlantic test set. Global foundation models exhibit a strong bias toward higher pressures (calmer weather), with Aurora suffering from severe pressure overestimation. CoDiCast hallucinates unphysically deep pressure minimums. GeoDES closely tracks the ground-truth distribution.}
    \label{fig:violin_slp}
\end{figure}

\begin{figure}[h]
    \centering
    \includegraphics[width=\textwidth]{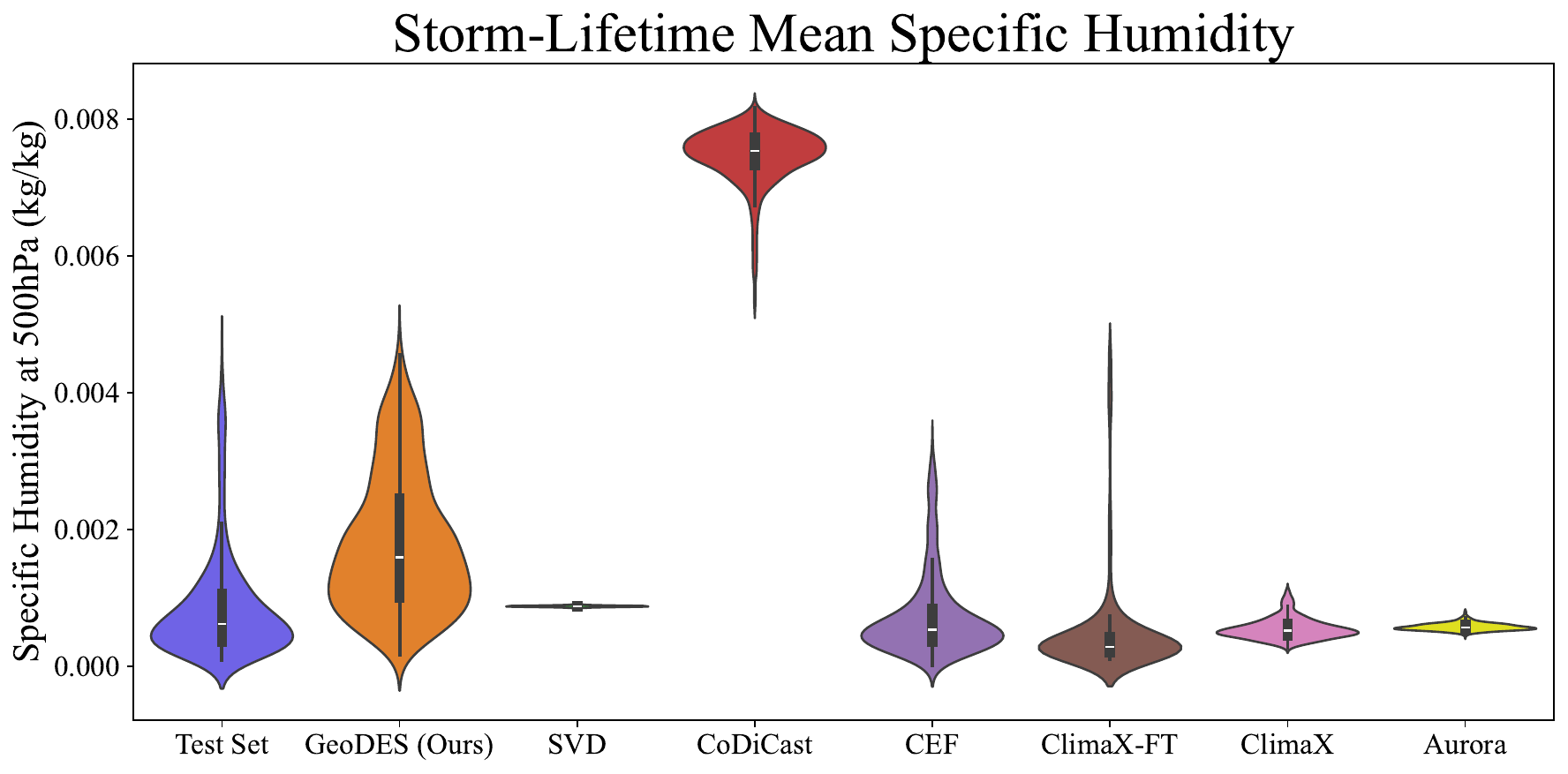}
    \caption{Distributions of storm-lifetime mean specific humidity at 500hPa across the North Atlantic test set. GeoDES captures the range of the true distribution, while most baseline models compress toward the calm-weather mean or generating unphysical outliers (CoDiCast).}
    \label{fig:violin_humidity}
\end{figure}

\textbf{Single-variable Storm-Lifetime Distributions:} We use violin plots to compare methods' storm-lifetime distributions of maximum wind speed in Figure~\ref{fig:violin_windmag}, minimum sea-level pressure in Figure~\ref{fig:violin_slp} and maximum humidity in Figure~\ref{fig:violin_humidity}.

Most methods fare acceptably with reproducing the general range and distribution of these single variables, except for the bias of global models towards calmer weather conditions (e.g. lower wind magnitudes, higher SLP and lower humidity). The two most notable exceptions are Aurora, with a strong bias towards high minimum SLP, and CoDiCast, with a bias towards low minimum SLP and apparent outlier values in its wind magnitude and humidity distributions. Despite both methods' use of batch-based normalization, these issues are perhaps artifacts of both methods' autoregressive sampling approaches. The core ($25\%-75\%$) percentiles of GeoDES' distributions track similarly to the test set distribution, although the violin plots represent a drawback of its normalization strategy: clipping between the $1$st and $99$th percentiles as described in Section~\ref{sec:method} prevents the model from generating the outlier values that naturally occur in the non-synthetic test set. Future work in this area could explore more advanced normalization strategies that enable the generation of reasonable outlier values without allowing the predictions to explode into the overly energetic failure modes demonstrated by CoDiCast or SVD.

% violin windmag

% violin slp

% violin humidity

\begin{figure}[htbp]
% \vspace{-0.6cm}
    \centering
    
    % --- First Row ---
    \begin{subfigure}[b]{0.24\textwidth}
        \centering
        \includegraphics[width=\textwidth]{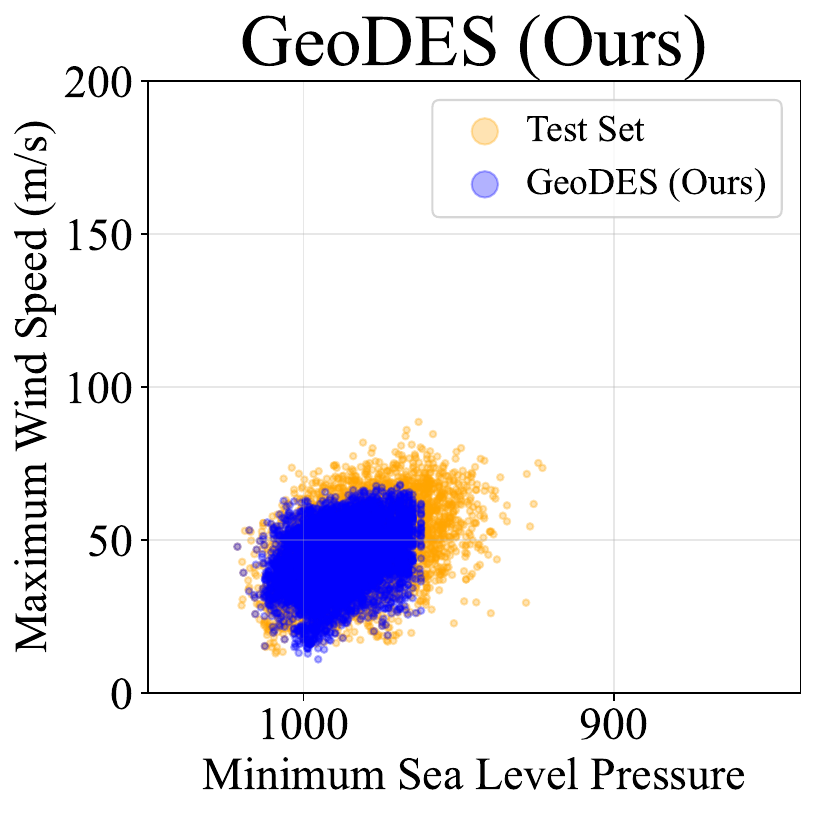}
        \label{fig:sub1}
    \end{subfigure}
    \hfill
    \begin{subfigure}[b]{0.24\textwidth}
        \centering
        \includegraphics[width=\textwidth]{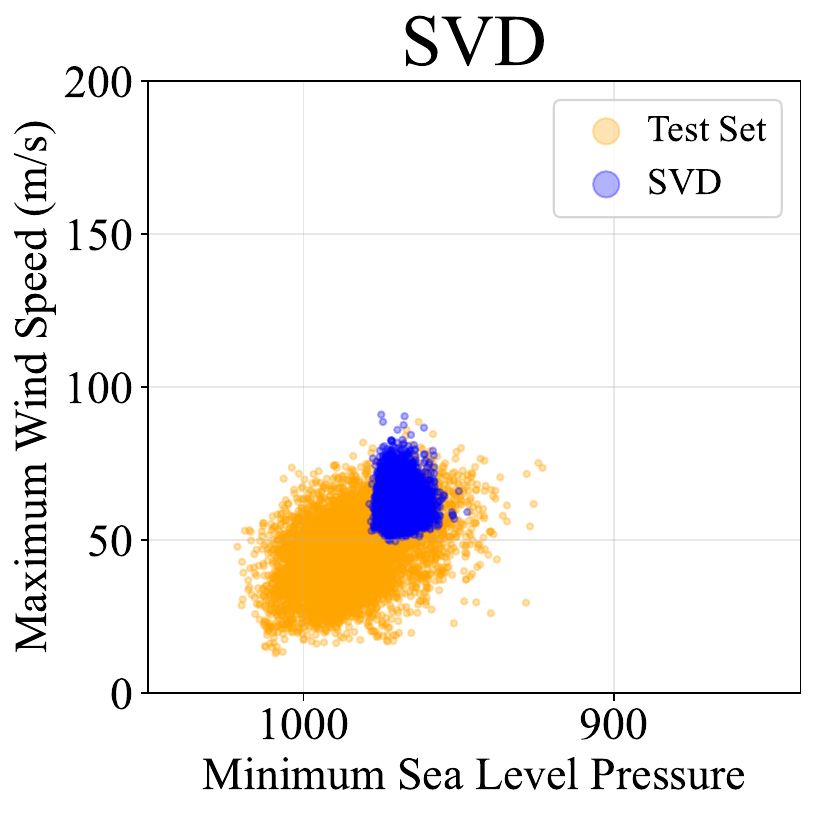}
        \label{fig:sub2}
    \end{subfigure}
    \hfill
    \begin{subfigure}[b]{0.24\textwidth}
        \centering
        \includegraphics[width=\textwidth]{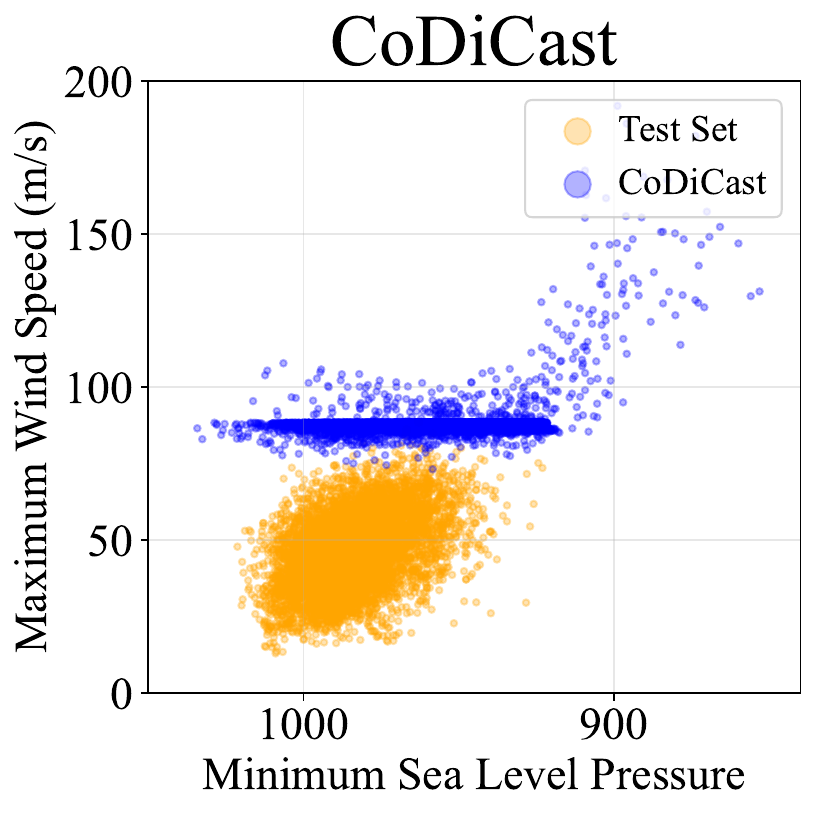}
        \label{fig:sub3}
    \end{subfigure}
    \hfill
    \begin{subfigure}[b]{0.24\textwidth}
        \centering
        \includegraphics[width=\textwidth]{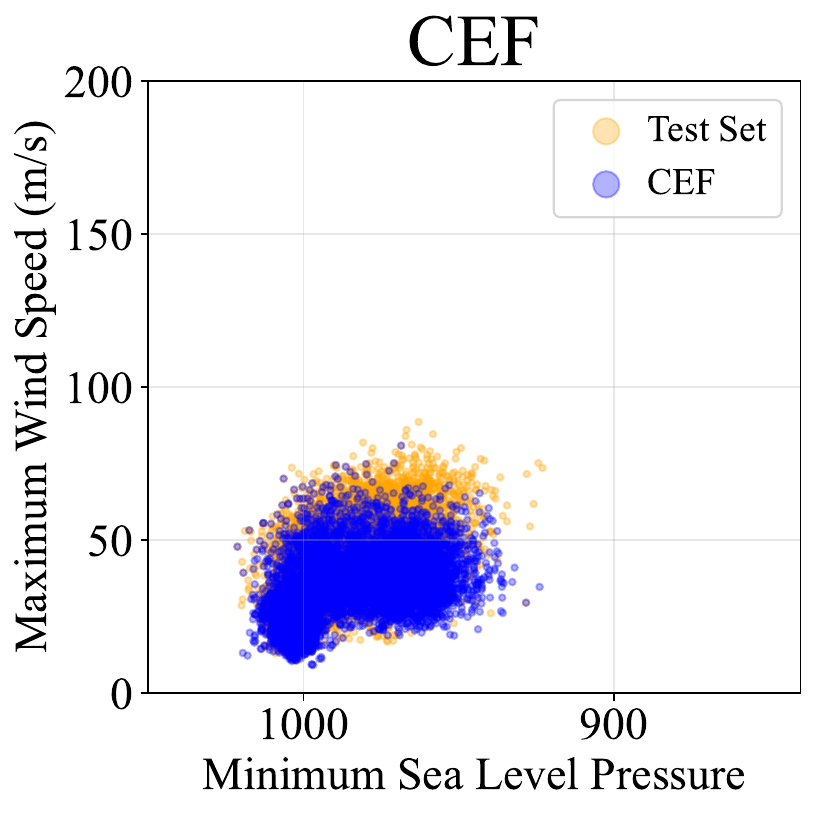}
        \label{fig:sub4}
    \end{subfigure}

    % --- Second Row ---
    \begin{subfigure}[b]{0.24\textwidth}
        \centering
        \includegraphics[width=\textwidth]{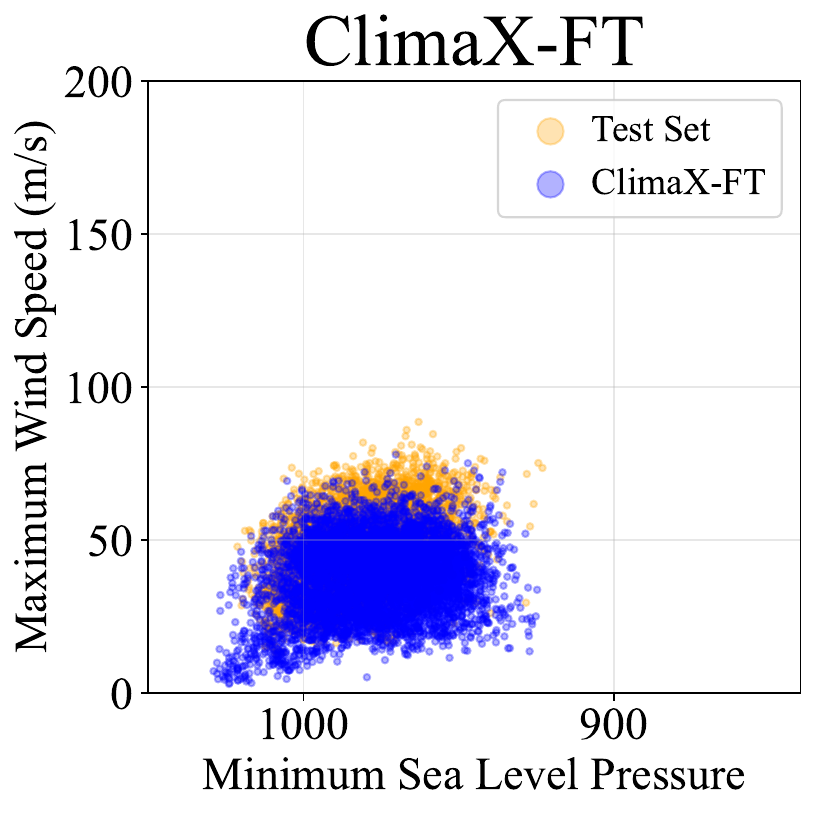}
        \label{fig:sub5}
    \end{subfigure}
    \hfill
    \begin{subfigure}[b]{0.24\textwidth}
        \centering
        \includegraphics[width=\textwidth]{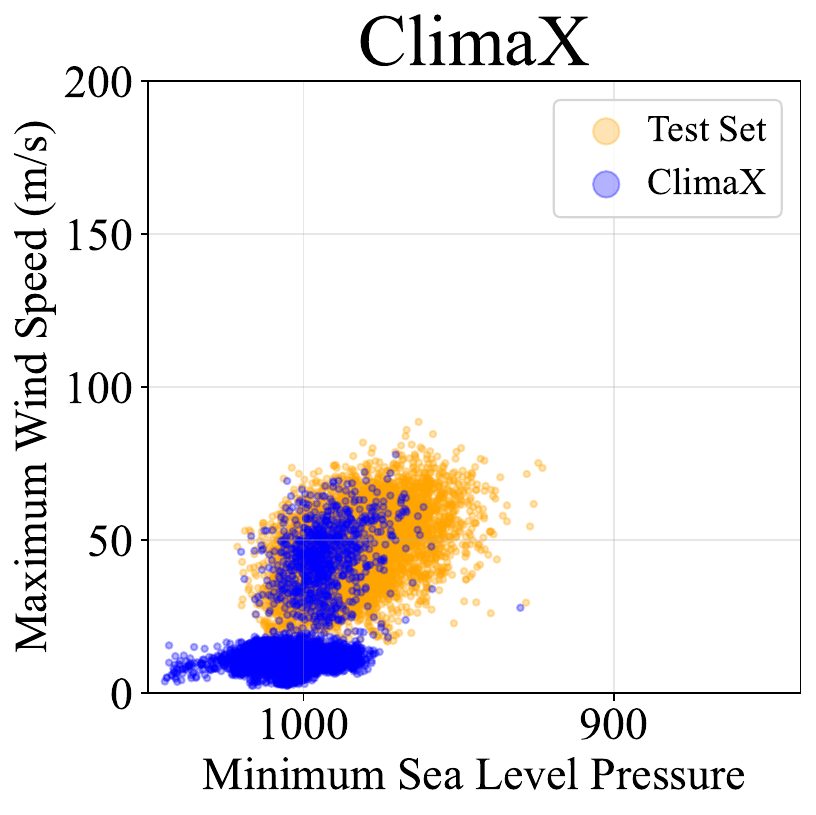}
        \label{fig:sub6}
    \end{subfigure}
    \hfill
    \begin{subfigure}[b]{0.24\textwidth}
        \centering
        \includegraphics[width=\textwidth]{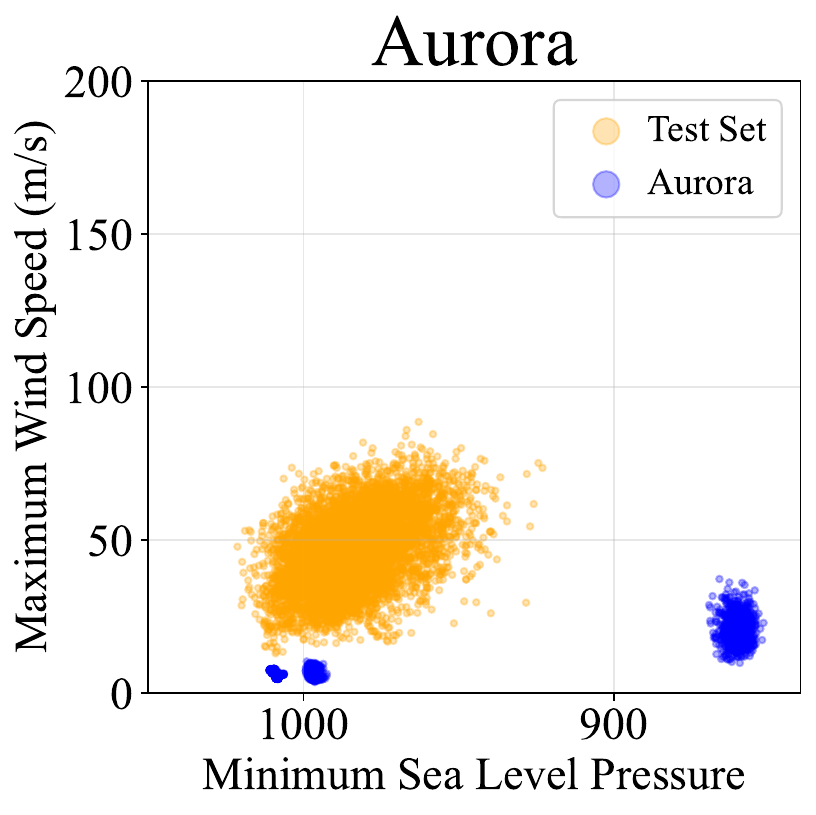}
        \label{fig:sub7}
    \end{subfigure}
    \hfill
    \begin{subfigure}[b]{0.24\textwidth}
        \centering
        \includegraphics[width=\textwidth]{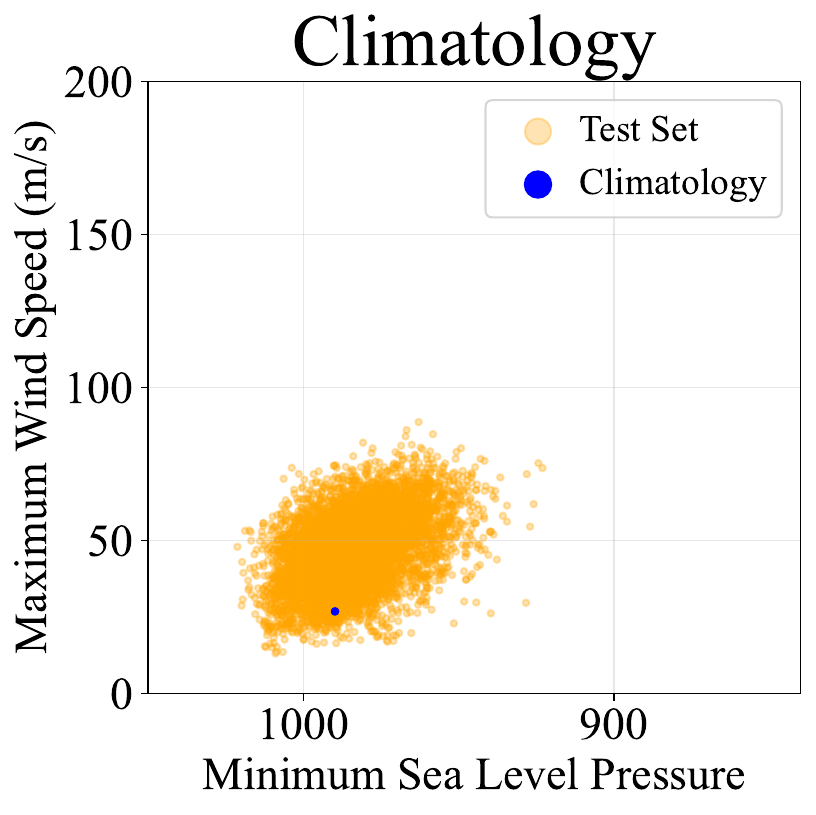}
        \label{fig:sub8}
    \end{subfigure}
% \vspace{-0.8cm}
\caption{The multivariate physical relationship between storm-lifetime maximum wind speed and minimum sea-level pressure. Each marker represents a single storm system. GeoDES (orange) best matches the ground-truth test set distribution (blue), demonstrating its ability to jointly model the low pressure storm core with the corresponding kinetic increase in wind speed. Global models systematically underestimate wind magnitude while overestimating pressure.} \label{fig:windmag-pressure}
\end{figure}

\textbf{Wind-Pressure Relationship:} Figure~\ref{fig:windmag-pressure} displays the relationships between storm-lifetime maximum wind speed (y-axis) and minimum sea-level pressure (x-axis) for each method, in orange, compared to the real test set, in blue. Each marker represents one storm. These plots are useful for discerning methods' multivariate relationship modeling capabilities.

\textit{GeoDES' wind-pressure relationship is most similar to that of the non-synthetic test set}. Similar to the patterns presented in previous metrics and plots, ClimaX and Aurora both underestimate wind magnitude and overestimate sea-level pressure in most storms. Oddly, Aurora's storms fall into three clear groups along the sea-level pressure axis (we perform one run, using the publicly available Aurora checkpoint and deterministic sampling; thus, these groups are not the result of differences across runs). ClimaX predictions fall into two general groups as well, with one group falling within the area occupied by the ground-truth test set storms and the other group falling into an area indicative of calm weather. ClimaX FT largely improves on these issues and resembles the test set more closely, although still with a bias towards calmer weather indicated by the large number of synthetic datapoints that fall below the area of test set points. A similar pattern is visible for the CEF model, while SVD and CoDiCast continue to output overly high-energy predictions.

% windmag-pressure relationship

\clearpage
\newpage
\subsection{Supplemental Metrics}\label{app:altmetrics}
We provide the following supplementary metrics for the synthesis experiments presented in Sections~\ref{sec:exp1}, \ref{sec:exp2} and \ref{sec:exp4}:
\begin{itemize}[topsep=0pt,itemsep=-0.01cm,leftmargin=*,partopsep=4pt, parsep=4pt]
\item \textit{Overall Root Mean Square Error (RMSE):} The standard pixel-wise RMSE across all un-normalized variables \citep{wilks2011statistical}.
\item \textit{Per-variable RMSE:} The standard pixel-wise RMSE calculated individually for each of the five atmospheric variables (sea-level pressure, U- and V-wind components, temperature and specific humidity) to identify variable-specific generative performance \citep{wilks2011statistical}.
\item \textit{Skill vs Persistence:} A comparative skill score evaluating the model's predictive accuracy relative to a naive persistence forecast (where the initial conditions at $t=0$ are held constant). It is calculated as $1 - (\text{MSE}_{\text{model}} / \text{MSE}_{\text{persistence}})$, where positive values indicate forecasting skill superior to persistence \citep{wilks2011statistical}.
\item \textit{Per-Variable Synoptic Anomaly Correlation Coefficient (SACC):} The spatial pattern correlation calculated individually for each atmospheric variable, following the application of a spatial Gaussian filter ($\sigma=1.5$) to mitigate the double-penalty effect for structurally sound but slightly displaced features \citep{jolliffe2012forecast}.
\item \textit{Anomaly Correlation Coefficient (ACC):} The standard spatial pattern correlation between the predicted and ground-truth anomaly fields without any spatial smoothing applied, providing a strict evaluation of localized structural alignment \citep{jolliffe2012forecast}.
\item \textit{Per-Variable Anomaly Correlation Coefficient (ACC):} The standard, un-smoothed ACC calculated individually for each atmospheric variable to isolate spatial correlation capabilities across disparate thermodynamic and kinematic fields \citep{jolliffe2012forecast}.
\end{itemize}

Results for the main experiment in Section~\ref{sec:exp1} are in Table~\ref{tab:exp1-supp}, Section~\ref{sec:exp2} on extreme storms in Table~\ref{tab:exp2-supp} and Section~\ref{sec:exp4} ablations in Table~\ref{tab:exp4-supp}. For each metric, best-performing models are bolded---as a naive baseline, we present climatology but do not bold its results. We analyze these results together, grouped by metric:

\textbf{RMSE and Skill:} As stated in Section~\ref{sec:evals-setup}, RMSE suffers from the ``double penalty'' problem where synthesis methods that generate realistic novel storms are penalized twice (for the presence of the storm features that did not occur in the ground truth storm, as well as the lack of storm features that did occur in the ground truth storm), whereas overly smoothed methods are only penalized for the lack of storm features occurring in the ground truth \citep{roberts2008scale, ebert2008fuzzy}. We include RMSE for completeness, due to the ubiquity of this metric.

\textit{GeoDES has the best overall synthesis model RMSE and skill vs persistence scores for both overall storm generation (Table~\ref{tab:exp1-supp}) and extreme storm generation (Table~\ref{tab:exp2-supp}).} Still, Climatology achieves the best scores by a wide margin---for example, GeoDES' general storms RMSE is $13.95$ compared to the Climatology's $9.79$ in Table~\ref{tab:exp1-supp}. This discrepancy highlights the known issues associated with this metric for weather synthesis assessment. Accordingly, it is difficult to assess whether a method scores low RMSE because it forecasts mostly accurate features and mostly in the same locations compared to the test set, or because it only predicts smooth predictions for all inputs to minimize its error. This pattern continues in the ablations, where the Linear Normalization method scores lowest RMSE in Table~\ref{tab:exp4-supp}. Because outlier values skew the linear normalization scheme and prevent the model from making effective use of its dynamic range, it is reasonable that the model learns to generate smooth outputs that are more similar to the climatology or global baselines such as ClimaX and Aurora.

\textbf{SACC and ACC:} Synoptic and non-synoptic ACC induce similar relative orderings of methods. Synoptic scores are globally higher than non-synoptic scores, indicating that all methods benefit from the use of the synoptic spatial filter. \textit{GeoDES scores the highest average ACC and SACC in overall storm generation, extreme storm generation and the ablations (Tables~\ref{tab:exp1-supp}, \ref{tab:exp2-supp} and \ref{tab:exp4-supp}).} Most global baselines perform poorly on these metrics, while Climatology is surprisingly competitive, with best overall (S)ACC in extreme storm generation (Table~\ref{tab:exp2-supp}). The storm-centric baseline (SVD) is the only method that is significantly aided by the synoptic spatial filter (e.g., its overall SACC score $0.15$ versus its ACC score of $0.03$ in Table~\ref{tab:exp1-supp}), but even this metric does not allow SVD to be competitive with other methods. \textit{SVD's poor performance highlights the importance of GeoDES' approach to achieving high performance as a storm-centered method, without the global context afforded to the other baselines.}

% supp metrics main
\begin{table}[]
\caption{Supplemental metrics to the North Atlantic storm synthesis experiment in Section~\ref{sec:exp1}. We report overall and per-variable Root Mean Square Error (RMSE), Skill vs. Persistence, Synoptic Anomaly Correlation Coefficient (SACC) and standard Anomaly Correlation Coefficient (ACC). GeoDES achieves the best overall generative performance among the evaluated dynamic models. Best results are bolded (excluding the statistical Climatology baseline).}
\label{tab:exp1-supp}
\resizebox{\textwidth}{!}{%
\begin{NiceTabular}{@{}r
>{\columncolor[HTML]{c3dcfe}}c cccccc
>{\columncolor[HTML]{EFEFEF}}c }
\toprule
\textbf{}                         & \textbf{GeoDES}                  & SVD                             & CoDiCast        & CEF                             & ClimaX FT                       & ClimaX                    & Aurora                    & \textit{Climatology} \\ \hline
Overall RMSE ($\downarrow$)       & $\mathbf{13.95\pm0.24}$ & $15.38\pm0.00$                  & $36.21\pm4.87$  & $16.10\pm0.24$                  & $18.19\pm1.38$                  & $16.50$                   & $14.47$                   & $\it 9.79$     \\
SLP RMSE ($\downarrow$)           & $15.27\pm1.46$                   & $15.73\pm0.00$                  & $35.74\pm5.07$  & $19.63\pm0.46$                  & $22.83\pm1.02$                  & $20.71$                   & \textbf{$\mathbf{14.02}$} & $\it 10.58$     \\
Wind U RMSE ($\downarrow$)        & $18.22\pm0.11$                   & $21.65\pm0.00$                  & $46.91\pm6.63$  & $17.84\pm0.45$                  & $19.53\pm1.32$                  & $\mathbf{17.15}$ & $18.22$                   & $\it 12.98$     \\
Wind V RMSE ($\downarrow$)        & $15.57\pm0.29$                   & $16.62\pm0.00$                  & $45.90\pm12.15$ & $20.10\pm1.04$                  & $17.39\pm0.10$                  & $\mathbf{14.32}$ & $15.16$                   & $\it 10.56$     \\
Temperature RMSE ($\downarrow$)   & $\mathbf{12.75\pm1.62}$ & $13.80\pm0.00$                  & $29.69\pm5.66$  & $13.71\pm0.15$                  & $21.09\pm3.83$                  & $20.80$                   & $17.00$                   & $\it 9.32$      \\
Humidity RMSE ($\downarrow$)      & $\mathbf{0.00\pm0.00}$  & $\mathbf{0.00\pm0.00}$ & $0.01\pm0.00$   & $\mathbf{0.00\pm0.00}$ & $\mathbf{0.00\pm0.00}$ & $\mathbf{0.00}$  & $\mathbf{0.00}$  & $\it 0.00$      \\
Skill vs Persistence ($\uparrow$) & $\mathbf{-0.66\pm0.03}$ & $-0.83\pm0.00$                  & $-3.30\pm0.58$  & $-0.91\pm0.03$                  & $-1.16\pm0.16$                  & $-0.96$                   & $-0.72$                   & $\it -0.16$     \\ \hline
Avg. SACC ($\uparrow$)            & $\mathbf{0.39\pm0.04}$           & $0.15\pm0.00$                   & $0.06\pm0.01$   & $0.19\pm0.03$                   & $0.28\pm0.00$                   & $0.09$                    & $0.10$                    & $\it 0.64$      \\
SLP SACC ($\uparrow$)             & $\mathbf{0.54\pm0.05}$  & $0.18\pm0.00$                   & $0.28\pm0.02$   & $0.38\pm0.07$                   & $0.33\pm0.04$                   & $0.15$                    & $0.44$                    & $\it 0.73$      \\
Wind U SACC ($\uparrow$)          & $0.61\pm0.04$                    & $0.18\pm0.00$                   & $-0.02\pm0.06$  & $0.35\pm0.09$                   & $\mathbf{0.64\pm0.02}$ & $0.14$                    & $0.08$                    & $\it 0.77$      \\
Wind V SACC ($\uparrow$)          & $\mathbf{0.48\pm0.02}$  & $0.12\pm0.00$                   & $0.00\pm0.02$   & $-0.31\pm0.15$                  & $-0.32\pm0.03$                  & $0.02$                    & $0.08$                    & $\it 0.71$      \\
Temperature SACC ($\uparrow$)     & $0.45\pm0.08$                    & $0.19\pm0.00$                   & $0.04\pm0.04$   & $0.44\pm0.09$                   & $\mathbf{0.51\pm0.02}$ & $-0.04$                   & $-0.11$                   & $\it 0.68$      \\
Humidity SACC ($\uparrow$)        & $\mathbf{0.10\pm0.01}$  & $0.04\pm0.00$                   & $-0.01\pm0.00$  & $0.07\pm0.04$                   & $0.02\pm0.05$                   & $0.00$                    & $0.04$                    & $\it 0.27$     \\ \hline
Avg. ACC ($\uparrow$)             & $\mathbf{0.37\pm0.04}$           & $0.03\pm0.00$                   & $0.06\pm0.01$   & $0.18\pm0.03$                   & $0.27\pm0.00$                   & $0.08$                    & $0.08$                    & $\it 0.63$      \\
SLP ACC ($\uparrow$)              & $\mathbf{0.54\pm0.04}$  & $0.05\pm0.00$                   & $0.28\pm0.02$   & $0.38\pm0.07$                   & $0.33\pm0.04$                   & $0.13$                    & $0.42$                    & $\it 0.74$      \\
Wind U ACC ($\uparrow$)           & $0.59\pm0.04$                    & $0.04\pm0.00$                   & $-0.01\pm0.05$  & $0.33\pm0.09$                   & $\mathbf{0.61\pm0.02}$ & $0.12$                    & $0.05$                    & $\it 0.76$      \\
Wind V ACC ($\uparrow$)           & $\mathbf{0.45\pm0.02}$  & $0.03\pm0.00$                   & $0.00\pm0.01$   & $-0.29\pm0.15$                  & $-0.30\pm0.03$                  & $0.01$                    & $0.07$                    & $\it 0.69$      \\
Temperature ACC ($\uparrow$)      & $0.43\pm0.08$                    & $0.05\pm0.00$                   & $0.03\pm0.04$   & $0.42\pm0.09$                   & $\mathbf{0.49\pm0.02}$ & $-0.04$                   & $-0.10$                   & $\it 0.66$      \\
Humidity ACC ($\uparrow$)         & $\mathbf{0.09\pm0.01}$  & $0.01\pm0.00$                   & $-0.01\pm0.00$  & $0.06\pm0.03$                   & $0.02\pm0.04$                   & $0.00$                    & $0.01$                    & $\it 0.25$ \\ \bottomrule
\end{NiceTabular}%
}
\end{table}

% supp metrics extreme
\begin{table}[]
\caption{Supplemental metrics to extreme result experiment in Section~\ref{sec:exp2}, where we evaluate on the top 10\% most severe storms in the North Atlantic test set by maximum wind magnitude. GeoDES maintains superior spatial pattern correlation (SACC and ACC) and the lowest overall RMSE among dynamic models even under extreme, out-of-distribution severity conditions.}
\label{tab:exp2-supp}
\resizebox{\textwidth}{!}{%
\begin{NiceTabular}{@{}r
>{\columncolor[HTML]{c3dcfe}}c cccccc
>{\columncolor[HTML]{EFEFEF}}c }
\toprule
                                   & \textbf{GeoDES}                  & SVD                             & CoDiCast        & CEF                             & ClimaX FT                       & ClimaX                   & Aurora                   & \textit{Climatology} \\ \hline
Overall RMSE ($\downarrow$)       & $\mathbf{14.79\pm1.31}$ & $17.50\pm0.00$                  & $37.43\pm5.93$  & $18.26\pm0.16$                  & $18.11\pm1.13$                  & $19.88$                  & $18.16$                  & $\it 11.69$     \\
SLP RMSE ($\downarrow$)           & $\mathbf{16.35\pm0.99}$ & $19.27\pm0.01$                  & $35.82\pm3.34$  & $18.77\pm0.60$                  & $21.01\pm2.11$                  & $27.13$                  & $19.97$         & $\it 14.20$     \\
Wind U RMSE ($\downarrow$)        & $\mathbf{20.88\pm2.42}$ & $25.94\pm0.00$                  & $52.01\pm11.78$ & $23.98\pm0.29$                  & $21.06\pm0.49$                  & $24.17$         & $25.90$                  & $\it 16.99$     \\
Wind V RMSE ($\downarrow$)        & $\mathbf{16.15\pm0.32}$ & $18.30\pm0.00$                  & $46.35\pm11.17$ & $22.30\pm1.13$                  & $19.82\pm0.18$                  & $17.20$         & $18.22$                  & $\it 12.47$     \\
Temperature RMSE ($\downarrow$)   & $\mathbf{11.20\pm2.66}$ & $12.35\pm0.00$                  & $28.11\pm4.44$  & $15.54\pm0.37$                  & $18.90\pm2.81$                  & $18.97$                  & $15.72$                  & $\it 6.13$      \\
Humidity RMSE ($\downarrow$)      & $\mathbf{0.00\pm0.00}$  & $\mathbf{0.00\pm0.00}$ & $0.01\pm0.00$   & $\mathbf{0.00\pm0.00}$ & $\mathbf{0.00\pm0.00}$ & $\mathbf{0.00}$ & $\mathbf{0.00}$ & $\it 0.00$      \\
Skill vs Persistence ($\uparrow$) & $\mathbf{-0.32\pm0.12}$ & $-0.56\pm0.00$                  & $-2.33\pm0.53$  & $-0.63\pm0.01$                  & $-0.61\pm0.10$                  & $-0.77$                  & $-0.62$                  & $\it -0.04$     \\ \hline
Avg. SACC ($\uparrow$)            & $\mathbf{0.45\pm0.06}$           & $0.16\pm0.00$                   & $0.06\pm0.02$   & $0.18\pm0.03$                   & $0.34\pm0.01$                   & $0.12$                   & $0.13$                   & $\it 0.70$      \\
SLP SACC ($\uparrow$)             & $\mathbf{0.58\pm0.10}$  & $0.21\pm0.00$                   & $0.29\pm0.03$   & $0.41\pm0.04$                   & $0.32\pm0.04$                   & $0.13$                   & $0.52$                   & $\it 0.79$      \\
Wind U SACC ($\uparrow$)          & $0.74\pm0.07$                    & $0.24\pm0.00$                   & $-0.04\pm0.08$  & $0.29\pm0.07$                   & $\mathbf{0.78\pm0.02}$ & $0.21$                   & $0.27$                   & $\it 0.90$      \\
Wind V SACC ($\uparrow$)          & $\mathbf{0.51\pm0.06}$  & $0.14\pm0.00$                   & $-0.03\pm0.04$  & $-0.30\pm0.19$                  & $-0.27\pm0.04$                  & $-0.01$                  & $0.04$                   & $\it 0.75$      \\
Temperature SACC ($\uparrow$)     & $0.55\pm0.12$                    & $0.22\pm0.00$                   & $0.01\pm0.08$   & $0.44\pm0.12$                   & $\mathbf{0.57\pm0.03}$ & $-0.08$                  & $-0.20$                  & $\it 0.79$      \\
Humidity SACC ($\uparrow$)        & $\mathbf{0.13\pm0.02}$  & $0.05\pm0.00$                   & $-0.02\pm0.03$  & $0.06\pm0.04$                   & $0.00\pm0.05$                   & $-0.01$                  & $0.06$                   & $\it 0.26$      \\ \hline
Avg. ACC ($\uparrow$)             & $\mathbf{0.43\pm0.06}$           & $0.04\pm0.00$                   & $0.06\pm0.02$   & $0.18\pm0.03$                   & $0.33\pm0.00$                   & $0.10$                   & $0.10$                   & $\it 0.69$      \\
SLP ACC ($\uparrow$)              & $\mathbf{0.58\pm0.09}$  & $0.05\pm0.00$                   & $0.29\pm0.02$   & $0.41\pm0.04$                   & $0.32\pm0.04$                   & $0.12$                   & $0.49$                   & $\it 0.79$      \\
Wind U ACC ($\uparrow$)           & $0.71\pm0.07$                    & $0.05\pm0.00$                   & $-0.04\pm0.08$  & $0.28\pm0.07$                   & $\mathbf{0.75\pm0.02}$ & $0.18$                   & $0.18$                   & $\it 0.88$      \\
Wind V ACC ($\uparrow$)           & $\mathbf{0.49\pm0.06}$  & $0.03\pm0.00$                   & $-0.03\pm0.04$  & $-0.29\pm0.19$                  & $-0.26\pm0.04$                  & $-0.01$                  & $0.03$                   & $\it 0.73$      \\
Temperature ACC ($\uparrow$)      & $0.52\pm0.12$                    & $0.05\pm0.00$                   & $0.01\pm0.07$   & $0.42\pm0.13$                   & $\mathbf{0.55\pm0.03}$ & $-0.08$                  & $-0.18$                  & $\it 0.77$      \\
Humidity ACC ($\uparrow$)         & $\mathbf{0.11\pm0.02}$  & $0.01\pm0.00$                   & $-0.02\pm0.03$  & $0.06\pm0.03$                   & $0.00\pm0.04$                   & $0.00$                   & $0.02$                   & $\it 0.24$\\ \bottomrule
\end{NiceTabular}%
}
\end{table}

% supp metrics ablations
\begin{table}[]
\caption{Supplemental metrics to ablations in Section~\ref{sec:exp4}. The table highlights the performance degradation associated with removing Signal-to-Noise Ratio (SNR) weighting, deterministic sampling (Non-Stoch.), using standard linear scaling (Linear Norm.) and removing the 2D pretraining phase (3D W/Out 2D). The Linear Normalization ablation achieves the lowest RMSE, mathematically demonstrating that lower RMSE often correlates with unphysical spatial smoothing and dynamic range collapse rather than structural realism.}
\label{tab:exp4-supp}
\begin{NiceTabular}{@{}r
>{\columncolor[HTML]{c3dcfe}}c cccc}
\toprule
                                  & \textbf{GeoDES}                 & No SNR                  & Non-Stoch.              & Linear Norm.            & 3D W/Out 2D            \\ \hline
Overall RMSE ($\downarrow$)       & $13.95\pm0.24$                  & $14.61\pm0.21$          & $14.19\pm0.07$          & $\mathbf{13.87\pm0.86}$ & $18.93\pm4.88$         \\
SLP RMSE ($\downarrow$)           & $15.27\pm1.46$                  & $17.56\pm1.42$          & $\mathbf{14.77\pm0.34}$ & $14.89\pm1.13$          & $26.02\pm9.33$         \\
Wind U RMSE ($\downarrow$)        & $18.22\pm0.11$                  & $17.72\pm0.82$          & $19.19\pm0.24$          & $\mathbf{17.51\pm0.79}$ & $24.62\pm8.26$         \\
Wind V RMSE ($\downarrow$)        & $15.57\pm0.29$                  & $\mathbf{14.72\pm0.16}$ & $15.13\pm0.03$          & $16.31\pm2.05$          & $15.59\pm0.06$         \\
Temperature RMSE ($\downarrow$)   & $\mathbf{12.75\pm1.62}$         & $15.02\pm1.54$          & $13.84\pm0.30$          & $12.89\pm0.32$          & $15.73\pm0.96$         \\
Humidity RMSE ($\downarrow$)      & $\mathbf{0.00\pm0.00}$          & $\mathbf{0.00\pm0.00}$  & $\mathbf{0.00\pm0.00}$  & $\mathbf{0.00\pm0.00}$  & $\mathbf{0.00\pm0.00}$ \\
Skill vs Persistence ($\uparrow$) & $-0.66\pm0.03$                  & $-0.74\pm0.02$          & $-0.69\pm0.01$          & $\mathbf{-0.65\pm0.10}$ & $-1.25\pm0.58$         \\ \hline
Avg. SACC ($\uparrow$)            & $\mathbf{0.39\pm0.04}$ & $0.36\pm0.06$           & $0.37\pm0.01$           & $0.37\pm0.01$           & $0.33\pm0.02$          \\
SLP SACC ($\uparrow$)             & $\mathbf{0.54\pm0.05}$          & $0.53\pm0.03$           & $0.47\pm0.02$           & $0.51\pm0.01$           & $0.34\pm0.18$          \\
Wind U SACC ($\uparrow$)          & $\mathbf{0.61\pm0.04}$          & $0.52\pm0.15$           & $0.50\pm0.00$           & $0.56\pm0.04$           & $0.59\pm0.08$          \\
Wind V SACC ($\uparrow$)          & $0.48\pm0.02$                   & $0.49\pm0.03$           & $0.44\pm0.02$           & $0.44\pm0.03$           & $\mathbf{0.54\pm0.10}$ \\
Temperature SACC ($\uparrow$)     & $0.45\pm0.08$                   & $0.41\pm0.01$           & $0.43\pm0.02$           & $\mathbf{0.48\pm0.02}$  & $0.37\pm0.13$          \\
Humidity SACC ($\uparrow$)        & $0.10\pm0.01$                   & $0.09\pm0.06$           & $0.10\pm0.01$           & $\mathbf{0.12\pm0.03}$  & $0.07\pm0.03$          \\ \hline
Avg. ACC ($\uparrow$)             & $\mathbf{0.37\pm0.04}$          & $0.35\pm0.05$           & $0.35\pm0.00$           & $0.35\pm0.01$           & $0.30\pm0.03$          \\
SLP ACC ($\uparrow$)              & $\mathbf{0.54\pm0.04}$          & $\mathbf{0.54\pm0.03}$           & $0.48\pm0.02$           & $0.52\pm0.01$           & $0.33\pm0.20$          \\
Wind U ACC ($\uparrow$)           & $\mathbf{0.59\pm0.04}$          & $0.50\pm0.14$           & $0.48\pm0.00$           & $0.54\pm0.04$           & $0.56\pm0.07$          \\
Wind V ACC ($\uparrow$)           & $0.45\pm0.02$                   & $0.46\pm0.03$           & $0.41\pm0.02$           & $0.42\pm0.03$           & $\mathbf{0.51\pm0.09}$ \\
Temperature ACC ($\uparrow$)      & $0.43\pm0.08$                   & $0.39\pm0.01$           & $0.41\pm0.02$           & $\mathbf{0.45\pm0.02}$  & $0.34\pm0.13$          \\
Humidity ACC ($\uparrow$)         & $0.09\pm0.01$                   & $0.08\pm0.05$           & $0.08\pm0.01$           & $\mathbf{0.10\pm0.02}$  & $0.06\pm0.03$        \\ \bottomrule
\end{NiceTabular}
\end{table}

\clearpage
\section{Biases Introduced Via Whole-World Pretraining and Performance Across Basins} \label{app:wholeworld}

\begin{table}[]
\caption{GeoDES results for each basin after 2D-phase pretraining on all basins, according to the setup in Appendix~\ref{app:wholeworld}. Model performance on fidelity metrics scales with the observational fidelity of the underlying basin data, rather than raw data volume, achieving highest ACC/SACC in the densely-observed Northern Hemisphere (North Pacific/Atlantic) and fuzzier predictions in the satellite-dependent Southern Hemisphere, despite greater overall data quantity and accordingly better energy-matching metrics (Frequency Bias Index, Peak Vorticity Error) in the Southern Hemisphere.}
\label{tab:bybasin} \centering
\begin{NiceTabular}{@{}rcccc}
\toprule
                      & N. Atlantic    & S. Atlantic & N. Pacific & S. Pacific \\ \midrule
\# Train              & $8589$         & $11262$     & $5782$     & $16358$    \\
\# Test               & $893$          & $1280$      & $614$      & $1917$     \\
Overall RMSE ($\downarrow$)          & $14.60$ & $15.05$     & $12.96$    & $14.63$    \\
Peak Vorticity Error ($\downarrow$)  & $5.10$  & $1.91$      & $4.92$     & $2.72$     \\
HF Spectral Ratio ($\cdot 1$)        & $0.71$  & $1.30$      & $1.04$     & $0.91$     \\
Frequency Bias Index ($\cdot 1$)       & $1.16$  & $1.04$      & $0.79$     & $1.17$     \\
Fractions Skill Score ($\uparrow$) & $0.77$  & $0.78$      & $0.66$     & $0.83$     \\
Synoptic ACC ($\uparrow$)          & $0.33$  & $0.36$      & $0.37$     & $0.34$     \\
Synoptic ACC T-Final ($\uparrow$)  & $0.38$  & $0.40$      & $0.40$     & $0.41$     \\
ACC ($\uparrow$)                   & $0.32$  & $0.34$      & $0.35$     & $0.33$     \\
ACC T-Final ($\uparrow$)           & $0.36$  & $0.38$      & $0.38$     & $0.39$     \\ \bottomrule
\end{NiceTabular}
\end{table}

\textbf{Setup:}
To evaluate the generalization capabilities of our diffusion architecture, we experimented with training on all storms from any of the four ocean basins at the 2D pretraining phase, followed by fine-tuning on a specific basin at the 3D fine-tuning phase. Notably, the distribution of storms in the pretraining dataset is highly asymmetric, following the order: North Pacific < North Atlantic < South Atlantic < South Pacific.

\textbf{Results:}
Table~\ref{tab:bybasin} displays an at first unusual relationship between data volume and model performance. GeoDES achieved its highest spatial accuracy and optimal variance in the North Pacific (Overall RMSE 12.96, Spectral Ratio 1.03), despite this basin representing the smallest fraction of the pretraining data. 

Rather than a failure of the architecture, this hemispheric split highlights a well-documented disparity in global meteorological observational fidelity. Ground-truth reanalysis in the Northern Hemisphere is aided by dense on-site observations and aircraft reconnaissance, providing more structurally precise, high-fidelity targets. In contrast, Southern Hemisphere reanalysis relies predominantly on smoother, satellite-derived estimates characterized by higher spatial uncertainty.

The model's performance directly reflects this information gap. The sharp, high-quality gradients in the Northern Hemisphere data allowed GeoDES to learn precise spatial and multivariate relationships during the 2D phase, requiring minimal correction during 3D specialization. The high volume of South Atlantic data---a basin characterized by high wind shear and unorganized convection---taught the model a chaotic structural prior, contributing to the prediction of  over-energized, high-frequency fields (Spectral Ratio of $1.30$) generated when fine-tuning for that domain. Ultimately, GeoDES scales not with raw data volume, but with the physical fidelity of the underlying observations.

\end{document}